%% file: main.tex
\documentclass{article}
\usepackage[OT1]{fontenc}
\usepackage{iclr2027_conference,times}
\setcitestyle{citesep={,},yysep={,}}
\input{math_commands}

\usepackage{amssymb,amsthm}
\usepackage{booktabs,multirow,array,tabularx}
\usepackage{microtype,placeins,needspace}
\usepackage{graphicx}
\usepackage[hidelinks]{hyperref}
\usepackage{etoolbox}
\input{arxiv_setup}
\makeatletter
\AtBeginDocument{%
  \ifdefined\pdfrunninglinkoff
    \patchcmd{\@outputpage}{\set@typeset@protect}
      {\pdfrunninglinkoff\set@typeset@protect}{}
      {\PackageError{foremem-links}{Cannot protect the page header}{Check the output routine.}}%
    \patchcmd{\@outputpage}{\box\@outputbox}
      {\pdfrunninglinkon\box\@outputbox\pdfrunninglinkoff}{}
      {\PackageError{foremem-links}{Cannot protect the page footer}{Check the output routine.}}%
    \pretocmd{\Hy@EveryPageBoxHook}{\pdfrunninglinkoff}{}
      {\PackageError{foremem-links}{Cannot protect the page anchor}{Check hyperref.}}%
  \else
    \PackageError{foremem-links}{pdfTeX 1.40.22 or newer is required}{Select a recent pdfLaTeX compiler.}%
  \fi
}
\AddToHook{shipout/background}[foremem-links]{\pdfrunninglinkoff}
\DeclareHookRule{shipout/background}{foremem-links}{before}{*}
\makeatother
\usepackage{url}
\usepackage{xcolor}

\newcommand{\Nbg}{N_{\mathrm{bg}}}
\newcommand{\ind}{\mathbf 1}
\newcommand{\norm}[1]{\left\lVert#1\right\rVert}
\input{current_results}

\title{Background Gradients Shape Memorization in Flow Matching}
\author{%
  \begin{tabular}{@{}c@{\hspace{1.5em}}c@{\hspace{1.5em}}c@{}}
    Xuanhua Yin\textsuperscript{1,*} &
    Boyu Wei\textsuperscript{1,*} &
    Shuyi Zhang\textsuperscript{2} \\
    {\scriptsize\texttt{xuanhua.yin@sydney.edu.au}} &
    {\scriptsize\texttt{bwei0951@sydney.edu.au}} &
    {\scriptsize\texttt{zhangshuyi2024@ia.ac.cn}} \\[0.6em]
    Shunqi Mao\textsuperscript{1} &
    Chuanzhi Xu\textsuperscript{1} &
    Weidong Cai\textsuperscript{1,\textdagger} \\
    {\scriptsize\texttt{smao7434@sydney.edu.au}} &
    {\scriptsize\texttt{chuanzhi.xu@sydney.edu.au}} &
    {\scriptsize\texttt{tom.cai@sydney.edu.au}}
  \end{tabular}\\[0.7em]
  {\small\textsuperscript{1}School of Computer Science, The University of Sydney}\\
  {\small\textsuperscript{2}Institute of Automation, Chinese Academy of Sciences}\\[0.3em]
  {\footnotesize\textsuperscript{*}Equal contribution.\quad
   \textsuperscript{\textdagger}Corresponding author.}
}
\hypersetup{%
  pdftitle={Background Gradients Shape Memorization in Flow Matching},
  pdfauthor={Xuanhua Yin, Boyu Wei, Shuyi Zhang, Shunqi Mao, Chuanzhi Xu, Weidong Cai},
  pdfsubject={Preprint}
}
\date{}

\begin{document}
\maketitle

\input{Chapters/abstract}
\input{Chapters/teaser}
\input{Chapters/introduction}
\input{Chapters/related_work}
\input{Chapters/methodology}
\input{Chapters/experiments}

\FloatBarrier
\input{Chapters/conclusion}

\label{page:main-end}
\medskip

\bibliographystyle{iclr2027_conference}
\bibliography{reference}
\clearpage
\appendix
\section*{Appendix}
\input{Chapters/natural_transfer_details}

\FloatBarrier
\input{Chapters/controlled_geometry_supplement}
\input{Chapters/appendix}
\input{Chapters/panel_prediction_supplement}
\FloatBarrier
\input{Chapters/bh_followup_results}
\FloatBarrier
\input{Chapters/literature_context}
\end{document}

%% file: math_commands.tex
\usepackage{amsmath,amsfonts,bm}

\def\eqref#1{equation~\ref{#1}}

\def\1{\bm{1}}

\DeclareMathAlphabet{\mathsfit}{\encodingdefault}{\sfdefault}{m}{sl}
\SetMathAlphabet{\mathsfit}{bold}{\encodingdefault}{\sfdefault}{bx}{n}

\newcommand{\R}{\mathbb{R}}



%% file: arxiv_setup.tex
\iclrfinalcopy
\renewcommand{\headrulewidth}{0pt}
\makeatletter
\renewcommand{\@maketitle}{%
  \vbox{\hsize\textwidth
    {\LARGE\scshape\@title\par}
    \vskip 1em
    {\centering\normalfont\@author\par}
    \vskip 1.5em
  }%
}
\makeatother

%% file: current_results.tex
\newcommand{\GoldLambda}{0.657}
\newcommand{\GoldLambdaLow}{0.610}
\newcommand{\GoldLambdaHigh}{0.703}
\newcommand{\GoldA}{3.498}
\newcommand{\GoldBAbs}{2.297}
\newcommand{\LpipsLambda}{1.043}
\newcommand{\LpipsLambdaLow}{0.912}
\newcommand{\LpipsLambdaHigh}{1.181}

%% file: Chapters/abstract.tex
\begin{abstract}
Repetition is closely associated with memorization in generative models, but how other training images affect the retention and copying of targets remains unclear.
We study this question in class-conditioned flow matching, where images outside the target set form the background.
At fixed target repetition and same-class background row count, replacing repeated same-class images with distinct images reduces the target extraction rate from 80.7\% to 18.0\%.
To explain this effect, we develop a paired-trajectory framework that isolates target-induced parameter displacement and the background gradient response to it.
This response has an exact path-integrated curvature representation, connecting background loss geometry to target learning.
Reciprocal response transfer between repeated and distinct backgrounds changes target retention and copying in both directions, establishing the response's causal role.
After target removal, the response correction parallel to the target-induced displacement preserves approximately 90\% of the copying effects of full response transfer.
Directly scaling the displacement also changes copying without further training.
The post-removal copying effects of reciprocal transfer are reproduced across datasets and architectures.
Together, these results identify the background gradient response as a mechanism through which same-class training data shape the retention of target learning and the reproduction of target images.
\end{abstract}

%% file: Chapters/teaser.tex
\begin{figure}[!ht]
\centering
\includegraphics[width=\linewidth]{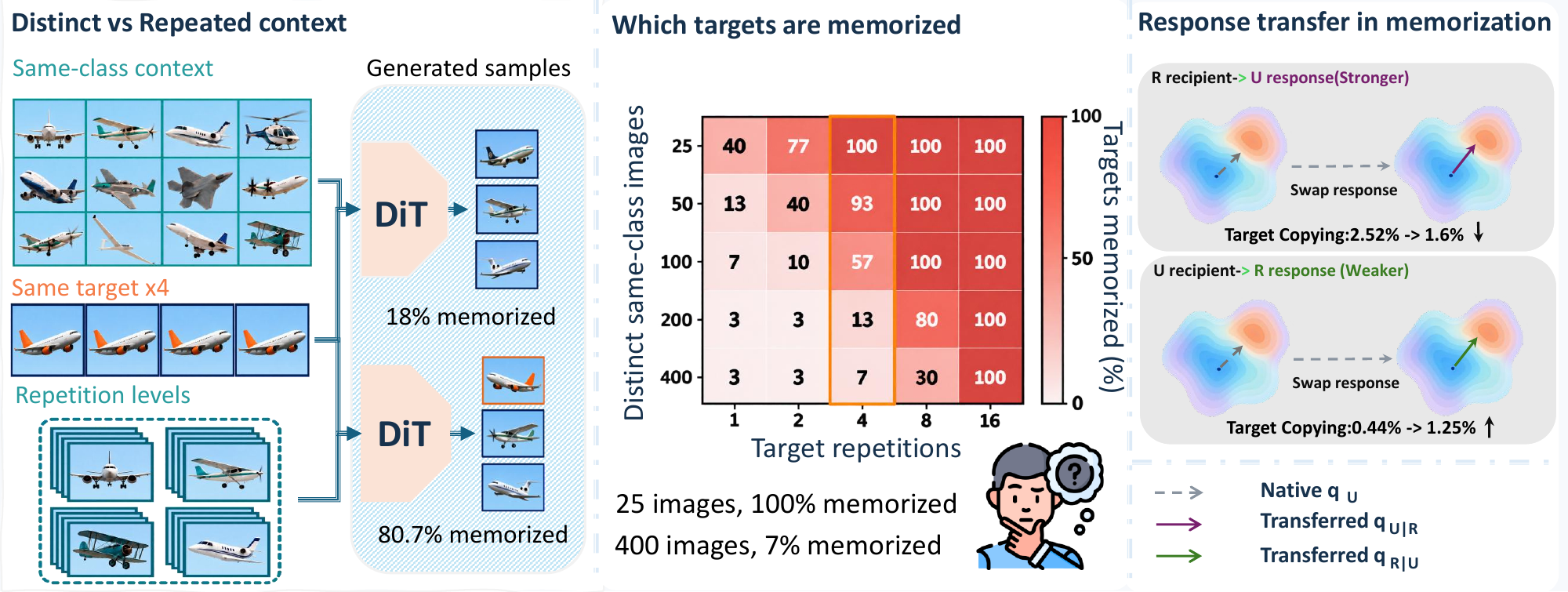}
\caption{Data composition and target learning. Left: at 400 same-class background
rows, repeated ($R$) and distinct ($U$) content give 80.7\% and 18.0\% target
extraction, pooled over target repetition counts. Pictured target copies are
schematic. Middle: balanced-design target extraction. Right: reciprocal
response transfer and interleaved per-generation target copying.}
\label{fig:teaser}
\end{figure}

%% file: Chapters/introduction.tex
\section{Introduction}
\label{sec:introduction}

Diffusion and flow-matching models can reproduce training images verbatim, a phenomenon known as memorization \citep{carlini2023extracting,somepalli2023diffusion,webster2023reproducible}. Empirical studies consistently find that an image is more likely to be memorized the more often it is repeated \citep{carlini2023extracting,somepalli2023understanding,ma2025invmm,gu2023memorization}, and deduplication has accordingly become a standard defense \citep{webster2023reproducible,kandpal2022dedup}. However, observed memorization varies widely across settings. Most images extracted in a Stable Diffusion study had at least a hundred near-duplicates \citep{carlini2023extracting}, whereas fivefold duplication increased inversion-based memorization on CelebAHQ-2.5k \citep{ma2025invmm}. These settings differ in model, training procedure, and detection method. We therefore investigate whether increasing the number of distinct images from a target's class reduces its memorization when target repetition is held fixed.

To test this possibility, a balanced 75-run design crosses target repetition with the number of distinct same-class images while holding total training-set size fixed. Images outside the target set form the background. As shown in the middle panel of Figure~\ref{fig:teaser}, expanding the background from 25 to 400 distinct same-class images reduces the target extraction rate (the fraction of targets detected at least once) from 100\% to 6.7\% at four target repetitions. A second experiment separates the number of distinct same-class background images from their training-row count. The left panel of Figure~\ref{fig:teaser} shows that, with that count fixed at 400, using 25 images repeated sixteen times yields 80.7\% target extraction, whereas using 400 distinct images yields 18.0\%. Repeating the same 25 background images instead of using each once changes extraction from 86.0\% to 80.7\%. The much larger change comes from replacing those images with distinct ones.

The mechanistic question is how training on these other images changes what the model retains from the targets. Answering it requires tracking the parameter changes induced by the targets, then testing how the background response to those changes affects retention and copying. To trace this composition effect through training, two trajectories start from the same model and optimizer state and use the same background data and stochastic inputs. Target updates are added to only one trajectory. The difference between their parameter states captures the displacement induced by the target set. The corresponding difference in background loss gradients defines the background gradient response. This response equals the background Hessian applied to the displacement, integrated along the line segment between the paired states. At a shared model state, the distinct background also has greater curvature along the initial target-gradient direction than the repeated background. The response measures background loss geometry along the displacement produced by the targets, complementing geometric accounts based on learned-manifold dimension and probability-landscape sharpness \citep{ross2025geometric,jeon2025sharpness}.

Reciprocal response transfer tests whether this response affects target retention and copying. A common target-learning phase gives the repeated and distinct conditions the same initial target-specific benefit. During continued training, each recipient retains its own background gradient at the parameter state without target updates. The alternative background supplies a replacement response evaluated at the recipient's current pair of states. When targets continue to appear at the same training steps across conditions, transferring the distinct-background response to the repeated-background model lowers the target copying rate (the fraction of generated images detected as target hits) from 2.520\% to 1.601\%. The reverse transfer raises copying in the distinct-background model from 0.439\% to 1.249\%.

The transfer effects exceed matched random-direction and preclip gradient-norm controls. They also persist when target updates stop: reciprocal transfer changes both the retained target-specific benefit and target copying after target removal. The post-removal copying effects reproduce in both directions on CINIC-10 with DiT and CIFAR-10 with a U-Net. Across the four primary transfers, conditional accuracy changes by less than 0.7 percentage points. Applying only the response correction parallel to the target-induced displacement preserves 90.8\% and 88.9\% of the full post-removal copying effects. Directly scaling this displacement also changes copying without further training, connecting target-induced parameter changes to generated copies.

Our contributions to understanding memorization in class-conditioned flow matching are threefold:
\begin{enumerate}
\item We disentangle the roles of distinct background content and repeated training rows in memorization. Controlled data interventions show that distinct same-class images reduce target extraction at fixed target repetition and background row count.

\item We develop a paired-trajectory framework that connects background loss geometry to target learning. It isolates the background gradient response to target-induced parameter displacement and gives this response an exact path-integrated curvature representation.

\item We establish a causal role for the background gradient response in target retention and copying. Reciprocal transfer and response-component interventions identify its effect, while displacement scaling connects target-induced parameter changes to generated copies. Post-removal transfer effects reproduce across the tested datasets and architectures.
\end{enumerate}

%% file: Chapters/related_work.tex
\section{Related Work}
\label{sec:related-work}

\paragraph{Data Composition and Memorization.}
Supervised-learning studies distinguish fitting arbitrary labels from the
role of rare examples in generalization
\citep{zhang2017rethinking,feldman2020longtail,feldman2020what}.
Diffusion-model studies link training-image replication to duplication
\citep{carlini2023extracting,somepalli2023understanding}, dataset size and
distribution \citep{gu2023memorization}, and local data coverage
\citep{merger2026local}. We isolate the effect of distinct same-class
content at fixed target repetition and background row count by replacing
repeated background images with distinct images, then investigate its
training-time mechanism in class-conditioned flow matching.

\paragraph{Geometry and Training Dynamics of Memorization.}
Geometric analyses characterize memorization through mismatched learned
and data-manifold dimensions \citep{ross2025geometric} and
probability-landscape sharpness \citep{jeon2025sharpness}. In flow matching,
the empirical optimal velocity field reproduces training samples,
directing attention to how learned models depart from this optimum
\citep{bertrand2025closed}. Training-time studies identify separate
timescales for generation and memorization \citep{bonnaire2025why},
learning-rate-induced regularization \citep{wu2025taking}, and forgetting
during subsequent training \citep{jagielski2023forgetting}. Complementary
analyses study the transition from memorization to generalization
\citep{kadkhodaie2024generalization,buchanan2025edge}. These accounts
leave unresolved how background loss geometry acts on target-induced
parameter changes. We identify this action through the finite background
gradient response between paired target-on and target-off states, which
equals the background Hessian action integrated along the segment
connecting them.

\paragraph{Training-Data Influence and Causal Interventions.}
Training-data attribution estimates example-level effects through
influence functions \citep{koh2017understanding}, checkpoint gradients
\citep{pruthi2020estimating}, and approximate optimization unrolling
\citep{bae2024unrolling}. Datamodels and scalable attribution relate
predictions to training-data choices \citep{ilyas2022datamodels,park2023trak},
with attribution methods also developed for diffusion models
\citep{zheng2024intriguing,mlodozeniec2025influence}. Memorization
interventions alter training through model aggregation and loss-based
filtering \citep{liu2024iterative}, or modify trained models through neuron
deactivation \citep{hintersdorf2024nemo} and geometry-aware parameter
sampling \citep{gegenfurtner2026reducing}. These methods do not test the
background response as a separately transferable component of training.
We isolate its effect on target retention and copying by reciprocally
exchanging full responses evaluated at the recipient's current paired
states while retaining its target-off raw background-gradient term.
Appendix~\ref{app:literature-context} develops the connections to data
duplication, optimization geometry, and generative-model evaluation
\citep{yin2026defaultshift,yin2026ship}.

%% file: Chapters/methodology.tex
\section{Methodology}
\label{sec:method}

At fixed target repetition, we test how same-class composition changes
memorization using paired trajectories, reciprocal response transfer,
response-component interventions, and direct displacement scaling.

Let $\mathcal T$ denote the target set, with all other training images
forming the background. Our primary experiments train a
class-conditional DiT \citep{peebles2023dit} on $32\times32$ RGB CIFAR-10
images \citep{krizhevsky2009cifar} with the flow-matching loss \citep{lipman2023flow}:
\begin{equation}
x_t=(1-t)\epsilon+tx,\qquad
\ell(\theta,x,c,t,\epsilon)
=\frac{1}{d}\left\|v_\theta(x_t,t,c)-(x-\epsilon)\right\|_2^2.
\label{eq:fm}
\end{equation}
Here $v_\theta$ is the class-conditioned velocity predictor with parameters
$\theta$, $x\in\mathbb R^d$ is an image with $d$ scalar coordinates and label
$c$, $\epsilon\sim\mathcal N(0,I_d)$, and $t\sim\mathrm{Uniform}[0,1]$.
The index $s$ denotes optimization steps.

\subsection{Controlled Data Interventions}
\label{sec:gold-method}

For a target image $x^\star$ of class $c$, let $\mathcal B_c$ contain
the other same-class images and $m(x)$ count occurrences of image $x$
in the training set. We distinguish target repetition, distinct same-class
image count, and other same-class training-row count as:
\begin{equation}
k=m(x^\star),\qquad
N_{\mathrm{bg}}=|\mathcal B_c|,\qquad
n_{\mathrm{bg}}^{\mathrm{row}}=\sum_{x\in\mathcal B_c}m(x).
\label{eq:data-factors}
\end{equation}
The count $k$ describes dataset multiplicity. Target draws depend on the
training sampler.

A balanced cyclic design evaluates each target under every combination
of $k$ and $N_{\mathrm{bg}}$, holding total target and background row counts
fixed across models and nesting image sets within a common pool.
We then fix both $k$ and
$n_{\mathrm{bg}}^{\mathrm{row}}$ to isolate the effect of distinct image
content. Comparing 400 different other same-class images with 25 images
repeated 16 times holds $n_{\mathrm{bg}}^{\mathrm{row}}=400$ fixed.
Comparing the latter with the same 25 images used once measures
the effect of repeated examples. An additional control replaces
these images with relabeled other-class images while retaining the
conditioning label.

For the mechanism experiments, each panel contains ten target images, one per class. Repeated
($R$), intermediate ($M$), and distinct ($U$) backgrounds use nested
sets of 25, 100, and 400 images per class, repeated 16, 4, and 1 times,
respectively. All three conditions preserve labels and background row
count. Their training objective is:
\begin{equation}
L_{\mathrm{mix}}^S(\theta)
=w_B L_B^S(\theta)+w_T L_T(\theta),\qquad S\in\{R,M,U\},
\label{eq:mixture}
\end{equation}
where $L_B^S$ and $L_T$ are mean losses on the background data and target
set. The fixed weights $w_B$ and $w_T$ equal their respective proportions
in the full training set. Stratified sampling preserves these weights, and
removing the target loss leaves $w_B$ unchanged. Numerical settings appear
in Section~4.1.

\subsection{Paired Training Trajectories and Background Gradient Responses}
\label{sec:paired-learning}
\label{sec:natural-response}

Figure~\ref{fig:response-overview} summarizes the design. A common target-learning
phase begins from a shared target-free checkpoint and produces the initial
target-specific benefit. The checkpoint supplies the model, optimizer, and random
states for target-on $(+)$ and target-off $(-)$ branches, which share background
data and stochastic inputs. Only the target-on branch receives the target loss.
At the phase endpoint $s_0$, we clone the paired parameters, optimizer states,
and random states across all backgrounds and intervention arms. Pulse-chase
training then removes targets, whereas interleaved training continues them on a
shared schedule with the same target sequence, flow times, and noise.

Let $g_B^S(\theta)=\nabla_\theta L_B^S(\theta)$ be the background loss
gradient. The target-induced parameter displacement and its finite
background gradient response are:
\begin{equation}
\delta\theta_{S,s}=\theta_{S,s}^{+}-\theta_{S,s}^{-},\qquad
q_{S,s}=g_B^S(\theta_{S,s}^{+})-g_B^S(\theta_{S,s}^{-}).
\label{eq:finite-response}
\end{equation}
The displacement includes direct target updates and subsequent trajectory
divergence, while the response measures its effect on the background gradient.
Both gradients use identical background examples, labels, flow times, and noise.

For a twice continuously differentiable background loss, write
$H_B^S(\theta)=\nabla_\theta^2L_B^S(\theta)$. The exact path-integral
response and, for $\delta\theta_{S,s}\ne0$, its finite-response projection are:
\begin{equation}
q_{S,s}=\int_0^1 H_B^S(\theta_{S,s}^{-}+\tau\delta\theta_{S,s})
\delta\theta_{S,s}\,d\tau,\qquad
\kappa_{S,s}=\frac{\delta\theta_{S,s}^{\top}q_{S,s}}
{\|\delta\theta_{S,s}\|_2^2}.
\label{eq:path-response}
\end{equation}
The projection equals the average directional curvature along the line
segment connecting paired parameters.
With $\xi_s$ denoting shared realized batch inputs, Equations~\ref{eq:finite-response}--
\ref{eq:path-response} hold for the batch loss, gradient, and Hessian defined in Appendix~\ref{app:mechanism-protocol}.

We additionally measure target-direction curvature at the shared target-free
checkpoint $\bar\theta$:
\begin{equation}
g_T=\nabla_\theta L_T(\bar\theta),\qquad
u=-\frac{g_T}{\|g_T\|_2},\qquad
h_S=u^\top H_B^S(\bar\theta)u.
\label{eq:pointwise-curvature}
\end{equation}
All backgrounds share $\bar\theta$, $u$, and the evaluation flow times and
noise. Hessian vector products use the full flow-matching loss. The
target-direction curvature $h_S$ compares background losses at a fixed
state and direction, whereas $\kappa_{S,s}$ measures the response along
the realized displacement.

\input{Chapters/method_overview}

\subsection{Reciprocal Response Transfer}
\label{sec:response-transfer}

To test the background response's effect on retention, we replace it while
retaining the recipient's baseline background gradient.
Let $A\in\{R,U\}$ denote the recipient background and
$C\in\{R,U\}$ the response source. At every step, we compute:
\begin{equation}
q_{C\mid A,s}
=g_B^C(\theta_{A,s}^{+})-g_B^C(\theta_{A,s}^{-}),\qquad
\widetilde g_{A\leftarrow C,s}^{+}
=g_B^A(\theta_{A,s}^{-})+q_{C\mid A,s}.
\label{eq:response-transfer}
\end{equation}
The source background loss is evaluated at the recipient's current paired
parameters. The baseline background gradient
$g_B^A(\theta_{A,s}^{-})$ is also recomputed at the current target-off
state. Setting $C=A$ recovers the native target-on background loss gradient.
The two interventions $R\leftarrow U$ and $U\leftarrow R$ replace the full
response vector. At the same recipient pair, the finite-response projection
$\kappa_{C\mid A,s}=\delta\theta_{A,s}^{\top}q_{C\mid A,s}/
\|\delta\theta_{A,s}\|_2^2$ is compared with the native projection
$\kappa_{A\mid A,s}$. Section~4.1 specifies their aggregation.

With $a_s$ indicating a scheduled target update, the complete target-on
gradient before clipping is:
\begin{equation}
G_{A\leftarrow C,s}^{+}
=w_B\widetilde g_{A\leftarrow C,s}^{+}
+a_s w_T\nabla_\theta L_T(\theta_{A,s}^{+}).
\label{eq:complete-transfer-gradient}
\end{equation}
For continued training after target removal, $a_s=0$. Interleaved training
retains the shared target schedule. The target-off branch uses
$w_Bg_B^A(\theta_{A,s}^{-})$.

The random-direction control replaces the weighted correction
$\Delta G_s=w_B(q_{C\mid A,s}-q_{A\mid A,s})$ with a per-step resampled
Gaussian direction, norm-matched within each parameter tensor. The
gradient-norm (preclip) control preserves the complete native gradient's
direction while matching the transferred gradient's global norm before clipping. Both use their own current recipient
pairs. After accumulation and unscaling, we construct and globally clip the
complete gradient, then apply AdamW with branch-specific momentum and
adaptive preconditioning (Appendix~A.1).

\paragraph{Response-Component Interventions.}
We decompose the source-to-native response correction along the recipient's
current target-induced displacement:
\begin{equation}
\begin{aligned}
\Delta q_s&=q_{C\mid A,s}-q_{A\mid A,s},
&d_s&=\frac{\delta\theta_{A,s}}{\|\delta\theta_{A,s}\|_2},\\
\Delta q_s^{\parallel}&=(d_s^\top\Delta q_s)d_s,
&\Delta q_s^{\perp}&=\Delta q_s-\Delta q_s^{\parallel}.
\end{aligned}
\label{eq:response-components}
\end{equation}
Here $\Delta q_s$ is the response correction and $d_s$ is the unit displacement
direction for $\delta\theta_{A,s}\ne0$. During post-removal training, we apply
$w_B[g_B^A(\theta_{A,s}^{+})+\Delta q_s^{\parallel}]$ or
$w_B[g_B^A(\theta_{A,s}^{+})+\Delta q_s^{\perp}]$, using the same clipping
and AdamW rule as full transfer. Each arm recomputes direction and correction
at its own current pair, retaining the complete native gradient. At a fixed pair,
$\Delta q_s^{\parallel}=(\kappa_{C\mid A,s}-\kappa_{A\mid A,s})\delta\theta_{A,s}$.
The parallel component isolates the finite-response projection change,
and the orthogonal component has zero displacement projection.
Finally, Section~\ref{sec:state-results} scales the terminal native interleaved
$R$ displacement, which jointly reflects the panel's ten targets, without
further training.

%% file: Chapters/method_overview.tex
\begin{figure}[t]
\centering
\includegraphics[width=\linewidth]{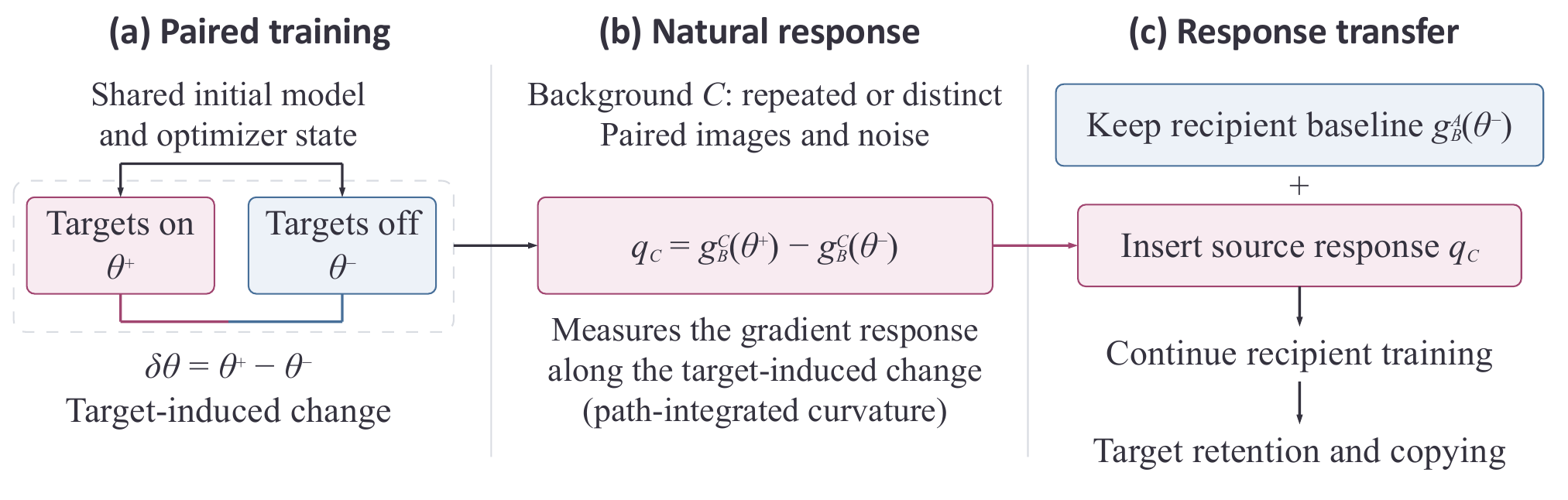}
\caption{Paired training and reciprocal response transfer. Target-on and
target-off states define the target-induced parameter displacement.
At the current recipient pair, an alternative background supplies the
response while the recipient retains its baseline background gradient.
Continuing training measures the resulting target retention and copying.}
\label{fig:response-overview}
\end{figure}

%% file: Chapters/experiments.tex
\section{Experiments and Results}
\label{sec:results}

\subsection{Evaluation and Statistical Setup}
\label{sec:evaluation-setup}
\label{sec:statistics}

The balanced design comprises 75 runs across three cycles,
with 30 targets each evaluated under all 25 combinations of target
repetition $k\in\{1,2,4,8,16\}$ and distinct background image count
$N_{\mathrm{bg}}\in\{25,50,100,200,400\}$.
Composition comparisons use 15 models across three blocks.
Primary data evaluations use step-10,000 checkpoints and 2,000
generated images per class and model, with EMA parameters for the
balanced design.

Primary response-transfer experiments use eight independent panels,
each containing ten targets, 4,000 background training rows, and
160 target training rows. Following 2,500 target-free warmup steps
and 400 common target-learning steps, paired continuations branch
at $s_0=2,900$. We evaluate both transfer directions after target
removal and under interleaved target training. At step 6,000,
we evaluate raw parameters using 4,096 generated images per class.
Training weights, sampling settings, and curvature-measurement
details appear in Appendix~\ref{app:mechanism-protocol}.

A generated image counts as a target hit when the target is its
nearest reference within the conditioning class and their distance
falls below the class-specific first percentile of holdout-to-reference
distances. Extraction rate is the proportion of targets with at least
one hit. Copying rate is the number of attributable target hits divided
by all generations of the class.
We use unnormalized pixel $\ell_2$ distance as the primary measure
and calibrate LPIPS-Alex \citep{zhang2018lpips} separately.
Conditional accuracy and KID \citep{binkowski2018mmd} accompany
both protocols. Reference sets, threshold calibration, and untrained
baselines are detailed in Appendix~\ref{app:measurement}.

To measure target-specific learning, we compare the paired target-on
and target-off models on each target $j$ and its untrained same-class
decoys $D_j$, using matched flow times and noise:
\begin{equation}
\Delta L_{S,X}(s)
= L_X(\theta_{S,s}^{-})-L_X(\theta_{S,s}^{+}),
\qquad X\in\{j,D_j\},
\label{eq:paired-loss}
\end{equation}
where $\theta_{S,s}^{+}$ and $\theta_{S,s}^{-}$ are the paired
parameters for continuation $S$, and $L_{D_j}$ averages the decoy
losses. Subtracting the decoy improvement gives the target-specific
benefit:
\begin{equation}
B_{S,j}^{\mathrm{spec}}(s)
= \Delta L_{S,j}(s)-\Delta L_{S,D_j}(s),
\qquad
\mathcal R_{S,j}(s)
= \frac{B_{S,j}^{\mathrm{spec}}(s)}
       {B_j^{\mathrm{spec}}(s_0)}.
\label{eq:retention}
\end{equation}
Each target's benefit at $s_0$ provides a fixed reference shared by
all continuations, so every continuation starts at
$\mathcal R_{S,j}(s_0)=1$.
We compute this ratio for each target before averaging within panels.
After target removal, $\mathcal R$ measures how much of the initial
target-specific benefit is retained. Under interleaved training,
it reflects both retention and continued target learning.

Statistical inference for mechanism experiments uses independent
panels. We first average outcomes across classes with equal weights
within each panel, then compute 95\% $t$ confidence intervals across
panels, using paired differences for intervention comparisons.
Within each training regime, Holm correction applies to four primary
copying comparisons: each of the two transfer directions versus
native training and versus its random-direction control.
Transfer fractions express mean copying effects relative to the
mean native $R$-to-$U$ copying gap in the same regime, with
paired-panel bootstrap intervals.
Projection averaging, additional inference procedures, and data-design
details appear in Appendices~\ref{app:mechanism-protocol},
\ref{app:mechanism-inference}, and~\ref{app:statistics}.

\subsection{Distinct Same-Class Images Reduce Target Extraction Rate}
\label{sec:data-results}
\label{sec:composition-results}

Every target appears under all 25 balanced-design combinations. Each model
contains 1,550 background and 62 target examples. At $k=4$, increasing
$N_{\mathrm{bg}}$ from 25 to 400 reduces target extraction rate from
100\% to 6.7\%, as shown in Figure~\ref{fig:teaser}.

\begin{figure}[t]
\centering
\includegraphics[width=\linewidth]{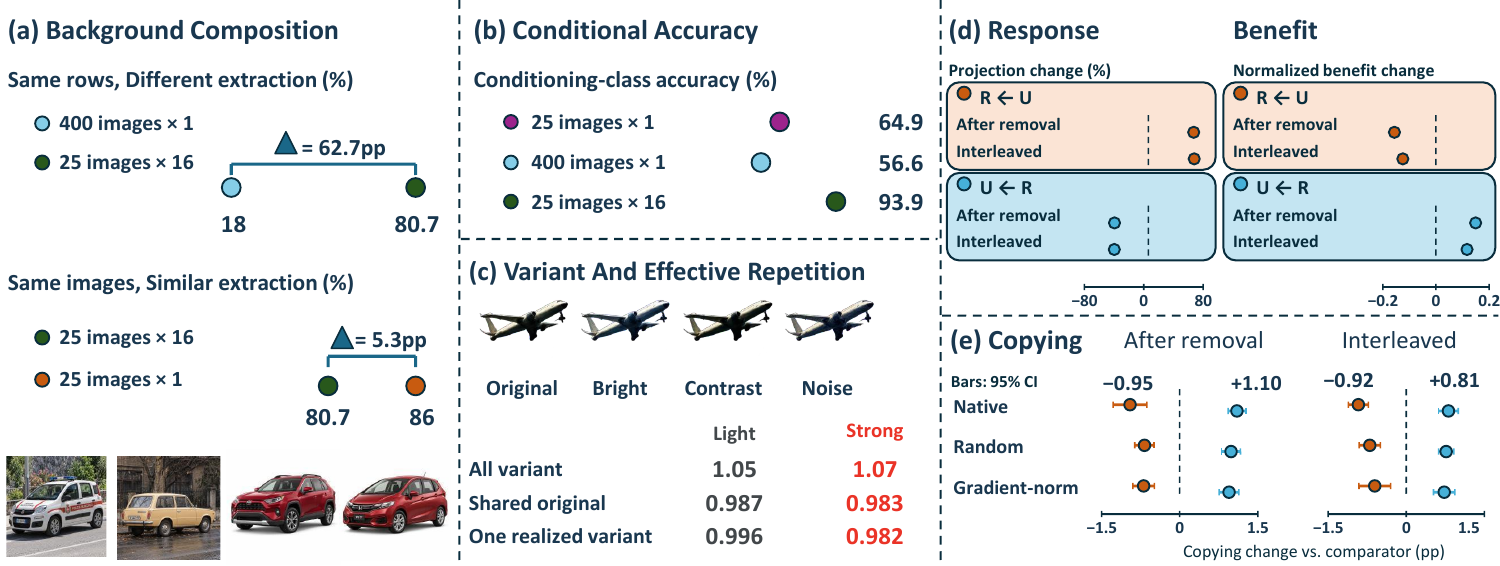}
\caption{Distinct background content reduces target extraction, and reciprocal
response transfer changes target-specific benefit and copying.
(a) Extraction under matched row or distinct-image counts.
(b) Corresponding conditional accuracy.
(c) Effective repetition of photometric variants relative to exact copies
under three detection-reference policies.
(d) Finite-response projection changes (\%) and normalized-benefit differences
from native values.
(e) Copying contrasts (transfer minus comparator, percentage points), with
paired 95\% confidence intervals across eight independent panels.
In (d) and (e), orange denotes $R\leftarrow U$ and blue denotes $U\leftarrow R$
after target removal or during interleaved training, where $R$ and $U$
use repeated and distinct backgrounds, respectively.}
\label{fig:results-overview}
\label{fig:response-transfer}
\end{figure}

Figure~\ref{fig:results-overview}a shows that, at fixed target repetition
and background row count, replacing 25 images
repeated 16 times with 400 distinct images reduces extraction from 80.7\%
to 18.0\%. Repeating the same 25 images instead of using them once changes
extraction from 86.0\% to 80.7\%. Distinct content produces a 62.7
percentage-point change, versus 5.3 points from repeating the same images.

At 20,000 steps, extraction remains lower with distinct images
(30.0\% versus 78.0\%), with conditional accuracies of 80.0\% and 94.0\%.
Figure~\ref{fig:results-overview}b shows the distinct condition's 56.6\%
accuracy at step 10,000. Its accuracy improves with longer training while
a 48.0-point extraction gap persists, as detailed in Table~\ref{tab:composition-budget}.

Figure~\ref{fig:results-overview}c shows that brightness, contrast, and
pixel-noise variants retain approximately the repetition strength of exact
copies when each target has one detection reference.
Appendix~\ref{app:near} specifies the photometric and reference-selection protocols.

\subsection{Background Composition Changes the Retention of Target Learning}
\label{sec:native-results}
\label{sec:natural-results}

All continuations share the same initial target-specific benefit at
$s_0=2,900$, with $\mathcal R(s_0)=1$.
After target removal, distinct backgrounds retain less of this benefit
and produce fewer target copies: terminal normalized benefit decreases
from 0.3517 under $R$ to 0.1428 under $U$, while copying decreases
from 2.685\% to 0.523\%.

As shown in Table~\ref{tab:native-chain}, both normalized target-specific
benefit and copying decrease across $R$, $M$, and $U$ under the shared
interleaved target schedule.
The $R$--$U$ copying gap is 2.081 percentage points
(95\% interval $[1.789,2.372]$).

These backgrounds also differ in their target-direction curvature
at the common target-free checkpoint.
Table~\ref{tab:native-chain} shows increasing curvature across $R$, $M$, and $U$
at fixed parameters and target direction, opposite to the ordering of
subsequent benefit and copying.
The $U$--$R$ curvature difference is 0.02665
(95\% interval $[0.02581,0.02748]$).
These geometric differences at a shared state motivate testing the
background gradient response through reciprocal transfer
in Section~\ref{sec:transfer-results}.

\begin{table}[!htbp]
\centering
\small
\setlength{\tabcolsep}{5pt}
\caption{Native backgrounds across eight experimental panels.
Target-direction curvature $h_S$ uses the shared step-2,500 checkpoint.
Normalized target-specific benefit and target copying rate use interleaved
training at step 6,000. Brackets give 95\% panel intervals.}
\label{tab:native-chain}
\begin{tabular}{@{}lrrr@{}}
\toprule
Background & $h_S$ & $\mathcal R$ & Target copying rate (\%)\\
\midrule
$R$, $25\times16$ & 0.03834 & 0.3206 & 2.520 [2.197, 2.843]\\
$M$, $100\times4$ & 0.05200 & 0.2197 & 1.177 [0.978, 1.376]\\
$U$, $400\times1$ & 0.06499 & 0.1305 & 0.439 [0.372, 0.506]\\
\bottomrule
\end{tabular}
\end{table}

\subsection{Reciprocal Response Transfer Changes Retention and Target Copying}
\label{sec:transfer-results}

Figure~\ref{fig:response-transfer}(d) and (e) shows that transferring the $U$ response
to $R$ lowers target-specific benefit and copying, whereas transferring the
$R$ response to $U$ raises both outcomes.

After target removal, transferring the $U$ response to $R$ increases the
recipient's finite-response projection by 67.3\%, lowers normalized
target-specific benefit from 0.3517 to 0.1954, and reduces copying from
2.685\% to 1.732\%. The reciprocal transfer decreases the recipient $U$
projection by 39.4\%, raises normalized benefit from 0.1428 to 0.2922,
and increases copying from 0.523\% to 1.622\%.

The same directional effects hold under interleaved target training.
Transferring the $U$ response to $R$ increases the projection by 68.0\%,
lowers normalized benefit from 0.3206 to 0.1959, and reduces copying
from 2.520\% to 1.601\%. Transferring the $R$ response to $U$ decreases
the projection by 39.7\%, raises normalized benefit from 0.1305 to 0.2475,
and increases copying from 0.439\% to 1.249\%.
These copying shifts amount to 44.2\% and 38.9\% of the native
$R$--$U$ copying gap, respectively.

As shown in Figure~\ref{fig:response-transfer}(e), the copying effects in both
directions exceed those of the random-direction
control and the gradient-norm control, which matches the transferred
gradient's norm before clipping while preserving the native gradient
direction.
Under interleaved training, $R\leftarrow U$ reduces copying by an additional
0.700 percentage points relative to its random-direction control
(95\% interval $[0.497,0.903]$), while $U\leftarrow R$ increases it by
an additional 0.766 points ($[0.620,0.912]$).
The four primary copying comparisons in each regime have Holm-adjusted
$p\le2.18\times10^{-4}$.
Conditional accuracy changes across the four primary transfers range
from $-0.524$ to $+0.688$ percentage points.

\paragraph{Parallel-Only Response Correction Preserves Most of the Copying Effect.}
Table~\ref{tab:components-main} compares the full, parallel-only, and
orthogonal-only corrections defined in Equation~\ref{eq:response-components}
after target removal.
The correction parallel to the target-induced displacement reduces copying
in $R$ by 0.865 percentage points and increases it in $U$ by 0.977 points.
The orthogonal correction changes copying in the same directions by
0.115 and 0.117 points.
The parallel effects exceed the orthogonal effects by
0.750 points (95\% interval $[0.725,0.775]$) in $R$ and
0.860 points ($[0.819,0.901]$) in $U$ across eight paired panels.
Parallel-only correction preserves 90.8\% and 88.9\% of the full-transfer
copying effects, respectively, locating most of the effect along the
target-induced displacement.

\input{Tables/response_components}

\input{Chapters/displacement_scaling}

\Needspace{5\baselineskip}
\subsection{Post-Removal Response Transfer across Datasets and Architectures}
\label{sec:replication-results}

Table~\ref{tab:cross-setting} shows both post-removal copying effects on
CINIC-10/DiT and CIFAR-10/U-Net, each with six independent panels,
matched background row counts, and random-direction controls.
$R\leftarrow U$ lowers copying and $U\leftarrow R$ raises it in both settings.
All four effects exceed their matched-random controls, with paired 95\%
intervals excluding zero.
Relative to the primary CIFAR-10/DiT experiments, these settings vary the
dataset or architecture separately, with targets absent throughout continuation.
Appendix~\ref{app:cross-setting-rates} provides further training details.

\input{Tables/cross_setting_main}

%% file: Tables/response_components.tex
\begin{table}[t]
\centering\small
\setlength{\tabcolsep}{8pt}
\caption{Response-component interventions after target removal. Copying rates
are percentages. Effects are native minus intervention for $R$ and intervention
minus native for $U$, in percentage points. Brackets are ordinary 95\% paired
$t$ intervals from the eight-panel component experiment.}
\label{tab:components-main}
\begin{tabular}{@{}lrrrr@{}}
\toprule
& \multicolumn{2}{c}{$R\leftarrow U$} & \multicolumn{2}{c}{$U\leftarrow R$}\\
\cmidrule(lr){2-3}\cmidrule(l){4-5}
Correction & Copying & Effect [95\% CI] & Copying & Effect [95\% CI]\\
\midrule
Native & 2.685 & n/a & 0.523 & n/a\\
Full & 1.732 & 0.953 [0.912, 0.994] & 1.622 & 1.099 [1.042, 1.156]\\
Parallel only & 1.820 & 0.865 [0.832, 0.898] & 1.500 & 0.977 [0.928, 1.026]\\
Orthogonal only & 2.570 & 0.115 [0.107, 0.123] & 0.640 & 0.117 [0.109, 0.125]\\
\bottomrule
\end{tabular}
\end{table}

%% file: Chapters/displacement_scaling.tex
\subsection{Target-Induced Parameter Displacement Controls Generated Copying}
\label{sec:state-intervention}
\label{sec:state-results}

We directly scale the joint ten-target displacement from the terminal native
interleaved $R$ pair without further training. At step 6,000, we evaluate:
\begin{equation}
\theta(\gamma)=\theta_{R,6000}^{-}+\gamma\delta\theta_{R,6000},
\qquad \gamma\in\{0,0.5,1,1.5\}.
\label{eq:state-intervention}
\end{equation}
The endpoints $\gamma=0$ and $1$ recover the target-off and target-on states.
Within each of the eight panels, scales share generation inputs and detection settings.
An equal-norm random parameter direction and a non-target learning direction
provide the controls reported in Table~\ref{tab:displacement-scaling-full}.

Copying rises from 0.33\% to 1.30\%, 2.520\%, and 3.71\% across the four
scales, increasing in 23 of 24 adjacent-scale comparisons within panels.
Target-specific benefit rises from zero to 0.172. The random and non-target
directions yield 0.38\% and 0.44\% copying, respectively. Decoy hits and
non-target copying remain within 0.30--0.31\% and 0.45--0.47\%.
From $\gamma=1$ to $1.5$, conditional accuracy changes by $-0.4$ percentage
points and KID by $+0.0011$. These state interventions connect the joint
target-induced displacement to generated copies, complementing the response
interventions that alter target learning during training.
Further implementation details appear in Appendix~\ref{app:displacement-scaling}.

\input{Tables/displacement_scaling_full}

%% file: Tables/displacement_scaling_full.tex
\begin{table}[!htbp]
\centering
\small
\setlength{\tabcolsep}{4pt}
\caption{State interventions on eight native interleaved $R$ pairs at step
6,000. Entries are panel means, and rates are percentages. $B^{\mathrm{spec}}$
uses a common target-off reference. The random direction matches the
$\gamma=1$ displacement norm.}
\label{tab:displacement-scaling-full}
\begin{tabularx}{\linewidth}{@{}Xrrrr@{}}
\toprule
State or direction & Target copy & $B^{\mathrm{spec}}$ & Decoy hit & Non-target copy\\
\midrule
$\gamma=0$ & 0.33 & 0 & 0.30 & 0.45\\
$\gamma=0.5$ & 1.30 & 0.071 & 0.30 & 0.45\\
$\gamma=1$ & 2.520 & 0.1284 & 0.30 & 0.46\\
$\gamma=1.5$ & 3.71 & 0.172 & 0.31 & 0.47\\
Random parameter direction & 0.38 & 0.004 & 0.30 & 0.45\\
Non-target learning direction & 0.44 & 0.006 & 0.31 & 0.46\\
\bottomrule
\end{tabularx}
\end{table}

%% file: Tables/cross_setting_main.tex
\begin{table}[!htbp]
\centering
\small
\setlength{\tabcolsep}{3pt}
\renewcommand{\arraystretch}{1.12}
\caption{Post-removal changes in target copying (percentage points).
Values are transfer minus comparator, with paired 95\% confidence intervals
across panels ($n=8$ for CIFAR-10/DiT and $n=6$ otherwise).
Random denotes the matched random-direction control.}
\label{tab:cross-setting}
\begin{tabular*}{\linewidth}{@{\extracolsep{\fill}}llr@{\hspace{5pt}}lr@{\hspace{5pt}}l@{}}
\toprule
\multirow{2}{*}{Setting} & \multirow{2}{*}{Comparator}
& \multicolumn{2}{c}{$R\leftarrow U$}
& \multicolumn{2}{c}{$U\leftarrow R$}\\
\cmidrule(lr){3-4}\cmidrule(l){5-6}
& & Change & \multicolumn{1}{c}{95\% CI}
  & Change & \multicolumn{1}{c}{95\% CI}\\
\midrule
\multirow{2}{*}{CIFAR-10 / DiT}
& Native & $\mathbf{-0.953}$ & $[-1.277,-0.630]$ & $\mathbf{+1.099}$ & $[0.929,1.269]$\\
& Random & $\mathbf{-0.676}$ & $[-0.863,-0.489]$ & $\mathbf{+0.986}$ & $[0.805,1.167]$\\
\addlinespace[6pt]
\multirow{2}{*}{CINIC-10 / DiT}
& Native & $\mathbf{-0.682}$ & $[-0.778,-0.586]$ & $\mathbf{+0.711}$ & $[0.558,0.864]$\\
& Random & $\mathbf{-0.584}$ & $[-0.716,-0.453]$ & $\mathbf{+0.615}$ & $[0.485,0.744]$\\
\addlinespace[6pt]
\multirow{2}{*}{CIFAR-10 / U-Net}
& Native & $\mathbf{-0.568}$ & $[-0.722,-0.414]$ & $\mathbf{+0.568}$ & $[0.494,0.642]$\\
& Random & $\mathbf{-0.488}$ & $[-0.590,-0.386]$ & $\mathbf{+0.503}$ & $[0.431,0.574]$\\
\bottomrule
\end{tabular*}
\end{table}

%% file: Chapters/conclusion.tex
\section{Conclusion}
We identify the background gradient response to target learning as a mechanism shaping memorization in class-conditioned flow matching. The paired-trajectory framework connects this response to background loss geometry. Reciprocal transfer establishes its causal role in target retention and copying. After target removal, the displacement-aligned correction preserves most of the effect of full transfer on copying. Scaling the target-induced displacement also changes copying without further training. Together, these findings connect background loss geometry to the retention of target learning and the reproduction of target examples.

Our mechanism experiments establish a causal role for the background response to jointly introduced targets in low-resolution, class-conditioned flow models. Isolating the response induced by individual targets would allow us to examine why the same background affects their retention differently at matched target exposure. Building on this analysis, future work could investigate which features of the relationship between targets and their background predict these retention differences, and whether those predictions also explain differences in generated copying.

%% file: Chapters/natural_transfer_details.tex
\section{Background Response, Target Benefit, and Generation Quality}
\label{app:natural-transfer-details}
The complete paired transfer comparisons report finite-response projections,
normalized target-specific benefit, copying, conditional accuracy, and KID.
Quality changes use the same receiver's native condition as the reference.
At the shared step-2,500 checkpoint, native directional Hessian values were
0.03834 [0.03784, 0.03884] for R, 0.05200 [0.05139, 0.05261] for M, and
0.06499 [0.06402, 0.06596] for U. Brackets give 95\% panel intervals.
Table~\ref{tab:transfer-curvature} separately reports the panel-mean relative
change in the finite-response projection, comparing source and native
background responses at the same recipient state pair. This diagnostic uses
the realized target-induced displacement, whereas $h_S$ uses the fixed direction $u$
at the common initial checkpoint. In the tables below, $\mathcal R$ measures
retention in pulse-chase and combines continued learning and retention in
interleaved training.

\begin{table}[htbp]
\centering\small
\setlength{\tabcolsep}{8pt}
\caption{Signed target-specific benefit and its normalized counterpart at step
6,000. The native and transferred arms use eight paired panels. Each
normalizer is fixed before continuation and shared across conditions for the
same panel-target. All four transfer comparisons agree in direction between
$B^{\mathrm{spec}}$ and $\mathcal R$ in all eight panels.}
\label{tab:signed-benefit}
\begin{tabular}{@{}llrr@{}}
\toprule
Regime & Condition & $B^{\mathrm{spec}}$ & $\mathcal R$\\
\midrule
Interleaved & R native & 0.1284 & 0.3206\\
 & M native & 0.0879 & 0.2197\\
 & U native & 0.0522 & 0.1305\\
 & R receiving U response & 0.0784 & 0.1959\\
 & U receiving R response & 0.0987 & 0.2475\\
\midrule
Pulse-chase & R native & 0.1412 & 0.3517\\
 & R receiving U response & 0.0782 & 0.1954\\
 & U native & 0.0571 & 0.1428\\
 & U receiving R response & 0.1171 & 0.2922\\
\bottomrule
\end{tabular}
\end{table}

\begin{table}[htbp]
\centering\small
\setlength{\tabcolsep}{9pt}
\caption{Native interleaved generation quality and LPIPS copying at step
6,000. Rates are percentages. KID uses unscaled Inception pool3 features.
The R row also describes the shared $\gamma=1$ dose endpoint.}
\label{tab:native-quality}
\begin{tabular}{@{}lrrr@{}}
\toprule
Background & LPIPS copy & Accuracy & KID\\
\midrule
R & 3.403 & 71.299 & 0.04121\\
M & 1.595 & 70.636 & 0.03899\\
U & 0.596 & 69.719 & 0.03741\\
\bottomrule
\end{tabular}
\end{table}

\paragraph{Inference Protocol.}
Each of the eight data panels is one independent unit. Classes are weighted
equally within a panel, while targets, generation banks, and generated draws
are treated as repeated measurements. Reported 95\% intervals are ordinary
$t$ intervals over the eight panel summaries. Contrast intervals use paired
panel differences. Within each regime, the four primary copying
contrasts (both transfer directions against native training and against the
corresponding matched-random control) form one Holm-adjusted family for
decisions. Transfer fractions divide the panel-mean intervention effect by the
native R-U copy gap in that regime and use paired panel bootstrap intervals.

\begin{table}[t]
\centering
\small
\setlength{\tabcolsep}{4pt}
\caption{Reciprocal response transfer and controls with full response replacement and eight paired panels per regime. The receiver supplies the training background, and the source supplies the response. Copy rates are percentages with 95\% panel intervals. Accuracy changes (pp) and KID changes are relative to the same receiver's native condition. The n/a entries mark native reference rows, for which changes are not reported. Random and norm controls use the corresponding transfer's matching constraints.}
\label{tab:response-transfer-full}
\begin{tabular}{@{}llrrrr@{}}
\toprule
Receiver & Intervention & $\mathcal R$ & Pixel copy [95\% CI] & $\Delta$Acc. & $\Delta$KID\\
\midrule
\multicolumn{6}{@{}l}{\emph{Pulse-chase}}\\
R & Native & 0.3517 & 2.685 [2.388, 2.981] & n/a & n/a\\
R & U response & 0.1954 & 1.732 [1.502, 1.961] & $-0.341$ & $-0.00128$\\
R & Matched random & 0.3146 & 2.408 [2.220, 2.596] & $-0.068$ & $-0.00054$\\
R & Preclip norm & 0.3157 & 2.424 [2.105, 2.743] & $-0.063$ & $-0.00038$\\
U & Native & 0.1428 & 0.523 [0.446, 0.600] & n/a & n/a\\
U & R response & 0.2922 & 1.622 [1.404, 1.840] & $+0.688$ & $+0.00172$\\
U & Matched random & 0.1539 & 0.636 [0.547, 0.724] & $+0.009$ & $-0.00010$\\
U & Preclip norm & 0.1669 & 0.674 [0.539, 0.809] & $+0.159$ & $+0.00036$\\
\midrule
\multicolumn{6}{@{}l}{\emph{Interleaved}}\\
R & Native & 0.3206 & 2.520 [2.197, 2.843] & n/a & n/a\\
R & U response & 0.1959 & 1.601 [1.323, 1.878] & $-0.524$ & $-0.00150$\\
R & Matched random & 0.3082 & 2.301 [1.915, 2.686] & $-0.130$ & $-0.00033$\\
R & Preclip norm & 0.2943 & 2.205 [1.839, 2.571] & $-0.096$ & $-0.00025$\\
U & Native & 0.1305 & 0.439 [0.372, 0.506] & n/a & n/a\\
U & R response & 0.2475 & 1.249 [1.037, 1.460] & $+0.530$ & $+0.00153$\\
U & Matched random & 0.1386 & 0.482 [0.375, 0.590] & $+0.245$ & $+0.00018$\\
U & Preclip norm & 0.1425 & 0.520 [0.447, 0.593] & $+0.118$ & $+0.00023$\\
\bottomrule
\end{tabular}
\end{table}

\begin{table}[t]
\centering
\small
\setlength{\tabcolsep}{6pt}
\caption{Finite-response projection changes (\%) under reciprocal transfer.
Source and native projections are measured at matched recipient states, then
averaged over four times and four batches per time within each panel before
taking their ratio (Appendix~\ref{app:mechanism-protocol}). Entries give means
and 95\% intervals across eight panels. Shared-checkpoint curvature $h_S$ is
reported separately in Table~\ref{tab:native-chain}.}
\label{tab:transfer-curvature}
\begin{tabular}{@{}llrr@{}}
\toprule
Receiver & Intervention & Pulse-chase & Interleaved\\
\midrule
R & U response & $+67.3$ [$+65.5$, $+69.0$] & $+68.0$ [$+65.3$, $+70.7$]\\
\midrule
U & R response & $-39.4$ [$-41.2$, $-37.6$] & $-39.7$ [$-40.9$, $-38.5$]\\
\bottomrule
\end{tabular}
\end{table}

\input{Chapters/mechanism_protocol_details}

\input{Chapters/mechanism_inference_details}
\input{Chapters/mechanism_paired_statistics}

\input{Chapters/additional_visual_summaries}

\input{Chapters/displacement_scaling_supplement}

%% file: Chapters/mechanism_protocol_details.tex
\subsection{Shared-Start Mechanism Protocol and Controls}
\label{app:mechanism-protocol}

Each panel uses an independent initialization seed and a fixed non-target
background pool for warmup through step 2,500. At this checkpoint, model,
optimizer, and random-number-generator states are cloned into target-on and
target-off branches. Common target learning occupies steps 2,500--2,900:
background data and stochastic inputs are paired, and only the on branch adds
the target loss. At $s_0=2{,}900$, both states, their optimizer states, and
their random-number-generator states are saved and cloned across $R$, $M$,
$U$, and the intervention continuations through step 6,000. Pulse-chase
continuations stop target training. Interleaved continuations retain the shared
target schedule. Each continuation subsequently maintains its own optimizer
state. Throughout, $w_B=4000/4160$ and $w_T=160/4160$ remain explicit.
Omitting the target term does not renormalize $w_B$. Stratified batches do not
replace these weights with sampling proportions. This protocol tests the
fate of a commonly acquired target benefit.

During common target learning and interleaved continuation, each stratified
batch contains 240 background draws and 16 target draws. The shared target
sampler seed gives the same target-ID sequence in every paired condition.
Table~\ref{tab:mechanism-exposure} reports actual per-target exposure rather
than exposure inferred from the dataset repetition count $k$, which denotes
the number of copied dataset rows. Conditions sharing a schedule have zero
difference in these counts. In pulse-chase, the target-update
indicator satisfies $a_s=0$ at every step from 2,901 through 6,000, yielding
zero target-training updates.

\begin{table}[!htbp]
\centering
\small
\setlength{\tabcolsep}{6pt}
\caption{Training schedule and actual target exposure. Counts are per target
in the target-on branch, whereas target-off branches omit the target loss.}
\label{tab:mechanism-exposure}
\begin{tabular}{@{}lrrr@{}}
\toprule
Phase & Step interval & Updates & Target draws\\
\midrule
Common target learning & 2,500--2,900 & 400 & 640\\
Interleaved continuation & 2,900--6,000 & 3,100 & 4,960\\
Pulse-chase continuation & 2,900--6,000 & 3,100 & 0\\
\bottomrule
\end{tabular}
\end{table}

For realized-batch notation, let $\xi_s$ collect the batch draws, flow times,
and noise at step $s$, and let $\widehat L_{B,s}^S(\theta,\xi_s)$ be the
corresponding background loss. Its gradient and Hessian are
$\widehat g_{B,s}^S=\nabla_\theta\widehat L_{B,s}^S$ and
$\widehat H_{B,s}^S=\nabla_\theta^2\widehat L_{B,s}^S$. The matched response is
$\widehat q_{S,s}=\widehat g_{B,s}^S(\theta_{S,s}^+,\xi_s)-\widehat
g_{B,s}^S(\theta_{S,s}^-,\xi_s)$. Because both states use the same $\xi_s$,
the finite-difference definition and path-integral identity in
Equations~\ref{eq:finite-response}--\ref{eq:path-response} hold for these batch
quantities after adding hats. The main text suppresses hats and $\xi_s$.

For each transfer or control branch, all gradient quantities are recomputed
at its current recipient pair using the same realized batch draws, flow times,
and noise at both states.
The complete gradients before clipping and the weighted transfer correction are:
\begin{align}
G_{\mathrm{nat}}
&=w_Bg_B^A(\theta_{A,s}^{+})
  +a_sw_T\nabla_\theta L_T(\theta_{A,s}^{+}),\\
\Delta G&=w_B\bigl(q_{C\mid A,s}-q_{A\mid A,s}\bigr),
\qquad G_{\mathrm{tr}}=G_{\mathrm{nat}}+\Delta G,
\label{eq:protocol-complete-gradients}
\end{align}
where $G_{\mathrm{nat}}$ and $G_{\mathrm{tr}}$ are the complete native and
transferred on-branch gradients, $\Delta G$ is their difference, $a_s$
indicates whether a target update is scheduled, $A$ is the recipient
background, and $C$ is the response source. The target-off branch
uses its weighted native background gradient. No source parameters or
optimizer moments are imported. The matched-random and preclip-norm controls
use:
\begin{align}
r_\ell&=\lVert\Delta G_\ell\rVert_2
\frac{z_\ell}{\lVert z_\ell\rVert_2},
&G_{\mathrm{random}}&=G_{\mathrm{nat}}+r,\\
G_{\mathrm{norm}}
&=\frac{\lVert G_{\mathrm{tr}}\rVert_2}
{\lVert G_{\mathrm{nat}}\rVert_2}G_{\mathrm{nat}},
\label{eq:protocol-gradient-controls}
\end{align}
where each parameter tensor forms one group $\ell$, including biases,
LayerNorm/adaLN parameters, and embeddings. The Gaussian vector $z_\ell$ is
redrawn each step from a seed independent of the training seed, and $r$
concatenates the group-wise corrections $r_\ell$. The resulting complete
gradients are
$G_{\mathrm{random}}$ and $G_{\mathrm{norm}}$. When
$\Delta G_\ell=0$, $r_\ell=0$. Zero-correction groups and other zero-norm
degeneracies are recorded separately. The norm control uses the global norm
over all parameters. Its scale factor is
$\alpha_s=\lVert G_{\mathrm{tr}}\rVert_2/\lVert G_{\mathrm{nat}}\rVert_2$.
In interleaved training it scales both the background and target terms.

Training uses bf16 autocast and fp32 master weights. After gradient
accumulation and unscaling, we compute $G_{\mathrm{nat}}$ and
$G_{\mathrm{tr}}$, construct the control gradient, clip its global norm at
1.0, and apply AdamW \citep{loshchilov2019adamw} with learning rate $10^{-4}$, $(\beta_1,\beta_2)=(0.9,0.999)$,
and zero weight decay. All branches use this order.
Table~\ref{tab:gradient-control-diagnostics} reports the pulse-chase matching
diagnostics. Matching preclip gradients does not match parameter updates
after momentum and adaptive scaling.

\begin{table}[!htbp]
\centering
\small
\setlength{\tabcolsep}{6pt}
\caption{Gradient-control diagnostics for reciprocal pulse-chase transfers.
Norm-ratio deviations are measured before clipping. IQR denotes the
interquartile range.}
\label{tab:gradient-control-diagnostics}
\begin{tabular}{@{}lrr@{}}
\toprule
Diagnostic & $R\leftarrow U$ & $U\leftarrow R$\\
\midrule
$\alpha_s$, median [IQR] & 1.08 [1.03, 1.14] & 0.93 [0.89, 0.97]\\
Maximum $\lVert G_{\mathrm{norm}}\rVert_2/\lVert G_{\mathrm{tr}}\rVert_2-1$
 & $3.1\times10^{-7}$ & $2.8\times10^{-7}$\\
Maximum $\lVert r_\ell\rVert_2/\lVert\Delta G_\ell\rVert_2-1$
 & $4.4\times10^{-7}$ & $4.0\times10^{-7}$\\
Zero-norm steps & 0 & 0\\
Zero-correction groups per step & 0 & 0\\
\bottomrule
\end{tabular}
\end{table}

For the common-checkpoint diagnostic, define:
\begin{equation}
g_T=\nabla_\theta L_T(\bar\theta),\qquad
u=-g_T/\lVert g_T\rVert_2,\qquad
h_S=u^\top H_B^S(\bar\theta)u,
\label{eq:protocol-hvp}
\end{equation}
where $\bar\theta$ is the target-free step-2,500 checkpoint and $L_T$ averages
the panel's ten targets. All parameter tensors are flattened in sorted-name
order. The direction spans all network parameters and is fixed before
downstream outcomes are observed. All three backgrounds share
$\bar\theta$, $u$, and the evaluation bank of flow times and noise. Target
images and stochastic inputs used to construct $u$ are also fixed. A
double-backward Hessian-vector product evaluates the full flow-matching
Hessian, including its residual term, without an added quadratic penalty.
Dot products accumulate in FP64 with TF32 and autocast disabled throughout
geometry measurement. Negative values are retained. The maximum deviation
$\lVert u\rVert_2-1$ was $2.2\times10^{-16}$, and all eight panels satisfied
$h_U>h_M>h_R$.

Retention uses the same decoy correction at the initial and continuation
steps:
\begin{align}
B_{S,j}^{\mathrm{spec}}(s)
&=\bigl[L_j(\theta_{S,s}^{-})-L_j(\theta_{S,s}^{+})\bigr]
 -\bigl[L_{D_j}(\theta_{S,s}^{-})-L_{D_j}(\theta_{S,s}^{+})\bigr],\\
\mathcal R_{S,j}(s)
&=B_{S,j}^{\mathrm{spec}}(s)/B_j^{\mathrm{spec}}(s_0),
\label{eq:protocol-retention}
\end{align}
where $j$ identifies a target, $D_j$ contains its fixed untrained same-class
decoys, and $B_j^{\mathrm{spec}}(s_0)$ is shared across continuations. Each
image loss averages 40 evaluations: eight fixed noise draws at each flow
time $t\in\{0.1,0.3,0.5,0.7,0.9\}$. Target and decoy losses use the same
times and noise, as does the target loss defining $u$.
Ratios are computed per target before panel averaging, with panels as the
statistical units. Signed,
unnormalized benefits are retained alongside ratios. The exclusion threshold
$|B_j^{\mathrm{spec}}(s_0)|<0.02$ was fixed before terminal outcomes were
examined and applied identically across conditions. Unnormalized benefits
retain any targets excluded from ratio summaries. Across the 200 panel-target
pairs in the three settings, all denominators were positive: their minimum,
median, and maximum were 0.304, 0.401, and 0.498. None was zero, negative, or
below the threshold, so all 200 pairs were included. For every pair, the
denominator difference across conditions was zero and
$\mathcal R_{S,j}(s_0)=1$.

Table~\ref{tab:signed-benefit} reports signed benefits and normalized
retention at step 6,000. The two summaries gave the same paired-effect
direction in all four reciprocal-transfer comparisons, with agreement in
all eight panels for each comparison. Pulse-chase ratios
measure retention after target training stops, whereas interleaved ratios combine
continued target learning and retention.

Both $B^{\mathrm{spec}}(s)$ and $\mathcal R(s)$ are recorded at
$s-s_0\in\{0,1,10,50,100,300,1000,3100\}$. Signed target-specific
benefit decreased at each sampled continuation time in both regimes,
as shown in Table~\ref{tab:benefit-trajectories}.

Finite-response comparisons use $\kappa_{C\mid A,s}$ at fixed recipient
states at steps $\mathcal S=\{3000,4000,5000,6000\}$, with four fixed random
batches at each step. For panel $p$, we first average over steps and batches:
\begin{equation}
\bar\kappa_{C\mid A}^{(p)}
=\frac{1}{16}\sum_{s\in\mathcal S}\sum_{b=1}^{4}
\kappa_{C\mid A,s,b}^{(p)},
\label{eq:protocol-response-aggregation}
\end{equation}
where $b$ indexes the evaluation batch. The panel's relative change is then
$100(\bar\kappa_{C\mid A}^{(p)}/\bar\kappa_{A\mid A}^{(p)}-1)\%$.
These panel changes are averaged to obtain the reported summary. The signed
absolute difference is
$\bar\kappa_{C\mid A}^{(p)}-\bar\kappa_{A\mid A}^{(p)}$.
Relative comparisons exclude a panel when
$|\bar\kappa_{A\mid A}^{(p)}|$ is below $10^{-3}$ times the overall median
native projection. No panel met this criterion. A random correction's
directional projection describes the effective correction and is not reported
as background Hessian curvature.

Generation evaluates raw step-6,000 parameters with 4,096 images per
class, pooled across two shared noise banks, using a 50-step Euler sampler.
The displacement-scaling evaluation uses the terminal native interleaved
$R$ pair with the same sampling settings.

\input{Tables/evaluation_designs}

%% file: Tables/evaluation_designs.tex
\begin{table}[t]
\centering
\small
\setlength{\tabcolsep}{4pt}
\caption{Evaluation settings. Generation budgets are per class and model.
Target extraction rate is an any-hit endpoint over target images, and target
copying rate is a per-generation endpoint. Photometric evaluation follows
Appendix~G.}
\label{tab:evaluation-designs}
\begin{tabularx}{\linewidth}{@{}lXrX@{}}
\toprule
Experiment & Evaluation point & Images & Endpoint\\
\midrule
Balanced & Step 10,000, EMA & 2,000 & Target extraction rate\\
Composition & Step 10,000 & 2,000 & Target extraction rate\\
Mechanism & Step 6,000, raw parameters & 4,096 & Target copying rate\\
\bottomrule
\end{tabularx}
\end{table}

%% file: Chapters/mechanism_inference_details.tex
\subsection{Additional Inference Details}
\label{app:mechanism-inference}
Balanced extraction outcomes are analyzed with bias-reduced logistic
regression. Target repetition and background diversity enter as
$\log_2(k/4)$ and $\log_2(N_{\mathrm{bg}}/100)$, with their fitted trade-off
summarized by $\lambda=-b_N/a$. Fieller intervals are used for this ratio, and
standard errors are clustered by target and run.

Within each training regime, the four primary copying contrasts (the two
transfer directions against native training and against their matched-random
controls) form one family for Holm-adjusted decisions. This familywise
adjustment is separate from the ordinary 95\% estimation intervals reported in
the tables. A transfer fraction divides the mean intervention effect by the
native R-U copying gap in the same regime. Its uncertainty is estimated by
paired bootstrap resampling over panels. The two additional pulse-chase
settings use six independent panels per setting and the same paired panel-level
summaries. Full transfer summaries and their inference protocol appear in
Appendix~\ref{app:natural-transfer-details}. Balanced-model specifications and
interval procedures appear in Appendix~\ref{app:statistics}.

%% file: Chapters/mechanism_paired_statistics.tex
\subsection{Paired Mechanism Contrasts}
\label{app:mechanism-paired-statistics}
All contrasts below first average the ten class-specific target measurements
within a panel, then compare paired arms across panels. Confidence intervals
use the two-sided Student-$t$ quantile with $n-1$ degrees of freedom, where
$n$ is the number of panels. Copy effects are measured in percentage points.
For receiver R, a positive effect is comparator minus transferred copying,
whereas for receiver U, it is transferred minus comparator copying. Thus, positive
effects represent reduction in R and increase in U. Individual intervals are
estimation intervals. The Holm adjustment applies to the tests, not the intervals.

For pulse-chase R, the random correction reduces copying by 0.277 pp
(95\% interval $[-0.062,0.617]$) relative to native training. Response transfer
reduces it by 0.953 pp and exceeds the random correction by 0.676 pp
($[0.489,0.863]$, with Holm-adjusted $p=1.21\times10^{-4}$).

\begin{table}[!htbp]
\centering\small
\setlength{\tabcolsep}{6pt}
\caption{Paired copying contrasts across eight panels in each primary regime.
The four native/random comparisons constitute one Holm family per regime.
Preclip comparisons are supplementary and are not assigned a $p$ value from
that four-test family (n/a).}
\label{tab:mechanism-paired-copy}
\begin{tabular}{@{}llrr@{}}
\toprule
Regime & Receiver / comparator & Effect [95\% CI] & Holm $p$\\
\midrule
Pulse-chase & R / native & 0.953 [0.630, 1.277] & $2.17\times10^{-4}$\\
 & U / native & 1.099 [0.929, 1.269] & $4.97\times10^{-6}$\\
 & R / random & 0.676 [0.489, 0.863] & $1.21\times10^{-4}$\\
 & U / random & 0.986 [0.805, 1.167] & $1.17\times10^{-5}$\\
 & R / preclip & 0.692 [0.486, 0.898] & n/a\\
 & U / preclip & 0.948 [0.758, 1.137] & n/a\\
\midrule
Interleaved & R / native & 0.919 [0.728, 1.111] & $2.76\times10^{-5}$\\
 & U / native & 0.810 [0.621, 0.998] & $3.85\times10^{-5}$\\
 & R / random & 0.700 [0.497, 0.903] & $8.16\times10^{-5}$\\
 & U / random & 0.766 [0.620, 0.912] & $2.03\times10^{-5}$\\
 & R / preclip & 0.605 [0.301, 0.908] & n/a\\
 & U / preclip & 0.729 [0.523, 0.934] & n/a\\
\bottomrule
\end{tabular}
\end{table}

Including both preclip contrasts in a supplementary six-test Holm family gives
adjusted $p$ values of $1.92\times10^{-4}$ and $2.78\times10^{-5}$ for
pulse-chase R and U, and $0.00218$ and $2.02\times10^{-4}$ for interleaved
R and U. This sensitivity analysis is separate from the four primary tests.

\begin{table}[!htbp]
\centering\small
\setlength{\tabcolsep}{7pt}
\caption{Paired differences in normalized target-specific benefit, with ordinary 95\% panel
intervals. Each entry uses eight panels and the target-wise normalization in
Equation~\ref{eq:retention}. The signs follow the copying contrasts: comparator
minus transfer for R, and transfer minus comparator for U.}
\label{tab:mechanism-paired-retention}
\begin{tabular}{@{}lrr@{}}
\toprule
Receiver / comparator & Pulse-chase & Interleaved\\
\midrule
R / native & 0.1562 [0.1298, 0.1827] & 0.1247 [0.1072, 0.1422]\\
U / native & 0.1494 [0.1413, 0.1575] & 0.1170 [0.0993, 0.1348]\\
R / random & 0.1191 [0.1005, 0.1378] & 0.1123 [0.0969, 0.1276]\\
U / random & 0.1383 [0.1264, 0.1503] & 0.1089 [0.0933, 0.1245]\\
R / preclip & 0.1203 [0.1005, 0.1401] & 0.0984 [0.0742, 0.1226]\\
U / preclip & 0.1253 [0.1074, 0.1433] & 0.1050 [0.0859, 0.1242]\\
\bottomrule
\end{tabular}
\end{table}

\paragraph{Transfer Fractions.}
The R and U transfer fractions are respectively 44.1\% [34.7, 51.7] and
50.8\% [44.0, 58.7] in pulse-chase, and 44.2\% [38.5, 49.6] and
38.9\% [35.3, 42.3] in interleaved training. Brackets are 95\% paired-panel
bootstrap intervals. The numerator is the panel-mean transfer effect and the
denominator is the panel-mean native R-U gap from the same regime. Panels are
resampled jointly in numerator and denominator. These are intervention effect
fractions, not natural mediation proportions.

\subsection{Complete Cross-Setting Copying Rates}
\label{app:cross-setting-rates}
The additional settings use six panels each with matched background row
counts and random-direction controls. The CIFAR-10/U-Net comparison retains
CIFAR-10 targets, flow matching, and AdamW. CINIC-10 is decoded-RGB
deduplicated. Table~\ref{tab:cross-setting-full}
reports native, transfer, and matched-random rates together with the paired
beyond-random effect. The primary DiT/CIFAR-10 arm rates are already reported
in Table~\ref{tab:response-transfer-full}. They are not additional replications.

\begin{table}[!htbp]
\centering\small
\setlength{\tabcolsep}{5pt}
\caption{Complete pulse-chase rates in the additional settings, each with six
panels. Native, transfer, and random columns give copying percentages with
95\% panel intervals. The last column gives the paired beyond-random effect
in percentage points. Settings are analyzed separately.}
\label{tab:cross-setting-full}
\begin{tabular}{@{}lrrrr@{}}
\toprule
Receiver & Native & Transfer & Random & Beyond random\\
\midrule
\multicolumn{5}{@{}l}{\emph{CINIC-10 / DiT}}\\
R & \shortstack{2.000\\{[1.606, 2.394]}} & \shortstack{1.318\\{[1.005, 1.630]}} & \shortstack{1.902\\{[1.516, 2.288]}} & \shortstack{0.584\\{[0.453, 0.716]}}\\\addlinespace
U & \shortstack{0.446\\{[0.347, 0.545]}} & \shortstack{1.157\\{[0.924, 1.389]}} & \shortstack{0.542\\{[0.426, 0.658]}} & \shortstack{0.615\\{[0.485, 0.744]}}\\\addlinespace
\midrule
\multicolumn{5}{@{}l}{\emph{CIFAR-10 / U-Net FM}}\\
R & \shortstack{1.742\\{[1.538, 1.947]}} & \shortstack{1.174\\{[1.086, 1.262]}} & \shortstack{1.662\\{[1.492, 1.832]}} & \shortstack{0.488\\{[0.386, 0.590]}}\\\addlinespace
U & \shortstack{0.401\\{[0.374, 0.428]}} & \shortstack{0.969\\{[0.897, 1.041]}} & \shortstack{0.466\\{[0.408, 0.524]}} & \shortstack{0.503\\{[0.431, 0.574]}}\\\addlinespace
\bottomrule
\end{tabular}
\end{table}

\input{Tables/cross_setting_gap}

%% file: Tables/cross_setting_gap.tex
\begin{table}[!htbp]
\centering
\small
\caption{Native background gaps during continued training after target
removal. Entries give the native $R$ minus $U$ target copying rate in
percentage points, with paired 95\% panel intervals. The primary setting
is included as the reference comparison.}
\label{tab:cross-setting-native-gap}
\begin{tabular}{@{}lrr@{}}
\toprule
Setting & Experimental panels & Native $R$ minus $U$\\
\midrule
CIFAR-10 with DiT, primary & 8 & 2.162 [1.884, 2.440]\\
CINIC-10 with DiT & 6 & 1.554 [1.249, 1.859]\\
CIFAR-10 with U-Net & 6 & 1.341 [1.143, 1.540]\\
\bottomrule
\end{tabular}
\end{table}

%% file: Chapters/additional_visual_summaries.tex
\clearpage
\subsection{Additional Visual Summaries}
\label{app:visual-summaries}
\begin{figure}[!htbp]
\centering
\includegraphics[width=\linewidth]{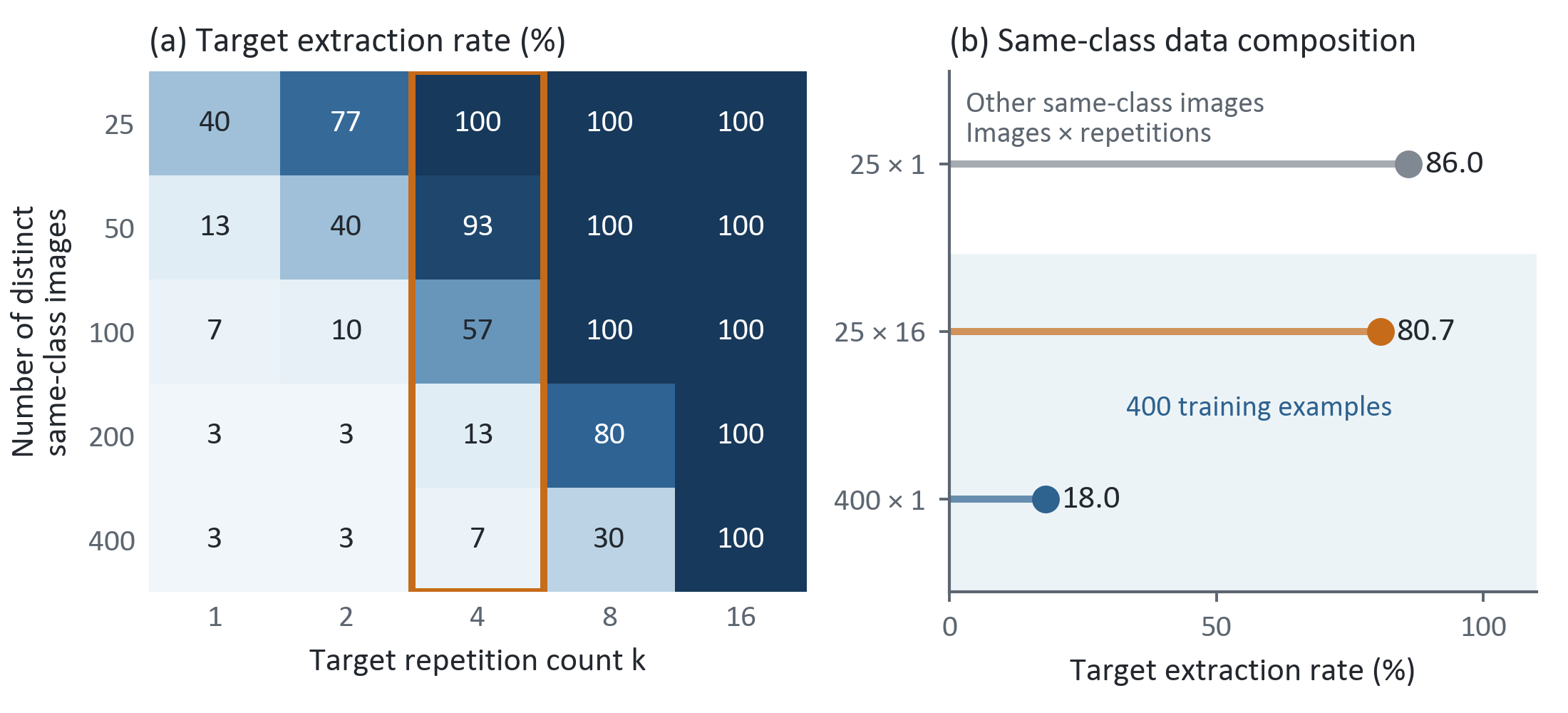}
\caption{Same-class data composition changes target extraction rate.
(a) The balanced design contains 30 target images per cell and 2,000
generated images per class and model. Cell labels are rounded percentages,
and the outline marks $k=4$. (b) Composition means over 150 paired target
evaluations per condition. Labels give distinct images multiplied by their
repetition counts. The shaded conditions contain 400 other same-class
training examples. Full coefficients and block analyses appear in Appendix~H.}
\label{fig:data-composition}
\label{fig:data-main}
\end{figure}

\begin{figure}[!htbp]
\centering
\includegraphics[width=\linewidth]{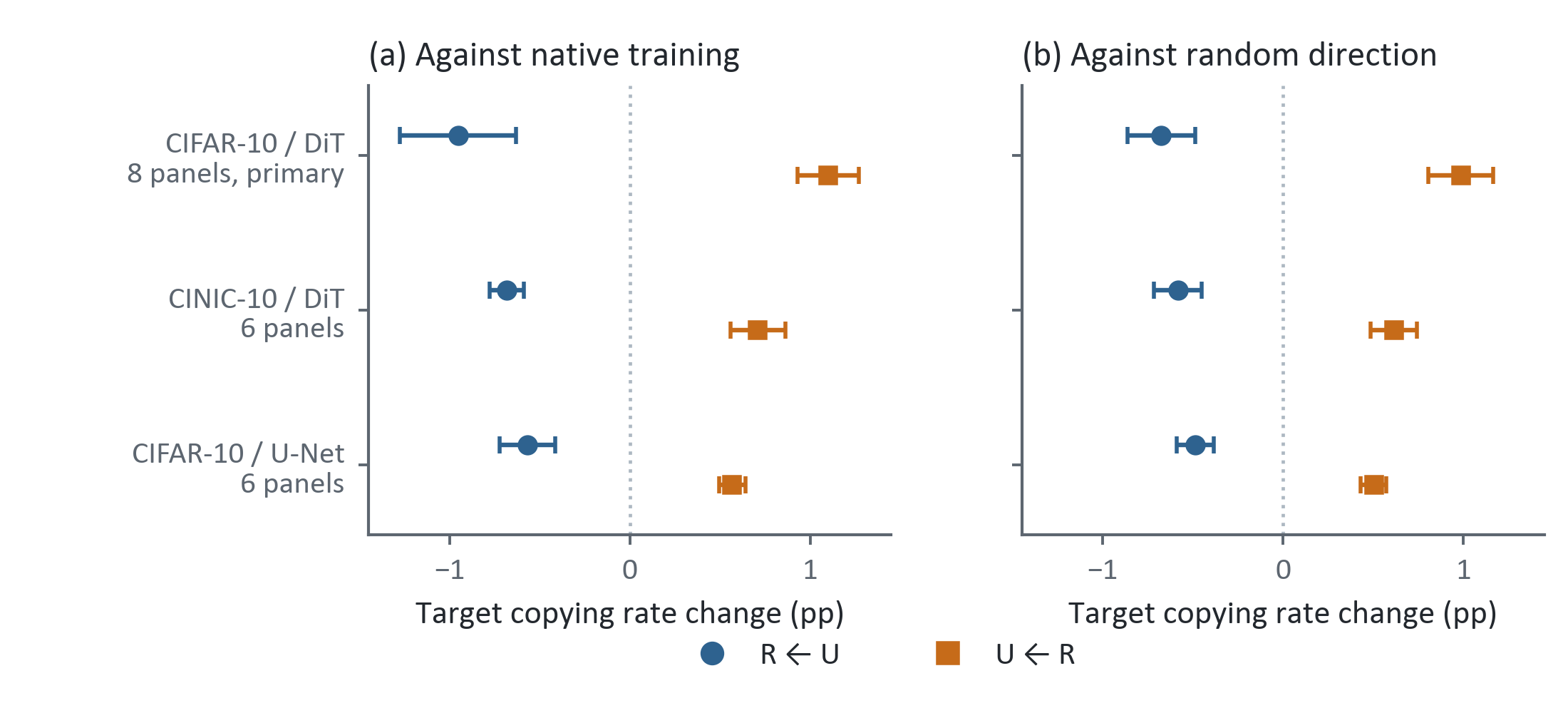}
\caption{Reciprocal response transfer across settings during continued
training after target removal. Changes in target copying rate are measured
against (a) native training and (b) random-direction controls. Negative
values indicate reduced copying under $R\leftarrow U$, and positive values
indicate increased copying under $U\leftarrow R$. Intervals are paired
95\% panel intervals. The primary CIFAR-10 with DiT result provides the
reference comparison with eight experimental panels. Each additional
setting uses six experimental panels. Native background gaps appear in
Table~\ref{tab:cross-setting-native-gap}.}
\label{fig:cross-setting}
\end{figure}

Figures~\ref{fig:data-composition} and~\ref{fig:cross-setting} visualize
the data-composition and cross-setting results. Numerical summaries appear
in Tables~\ref{tab:composition-full} and~\ref{tab:cross-setting}.

\paragraph{Relabeled-Background Control.}
Replacing 25 relabeled other-class images repeated 16 times with 400
distinct relabeled images reduces target extraction from 82.0\% to 20.0\%.
Their conditional accuracies are 5.8\% and 8.3\%, respectively.
These images retain the target conditioning label. Table~\ref{tab:composition-full}
reports the complete composition results.

\input{Tables/composition_full}

%% file: Tables/composition_full.tex
\begin{table}[t]
\centering
\small
\caption{Composition means over 150 paired target evaluations per condition.
Entries specify distinct images multiplied by repetition counts. Both rates
are percentages. Relabeled other-class images retain the target conditioning label.}
\label{tab:composition-full}
\begin{tabular}{@{}lrr@{}}
\toprule
Background composition & Target extraction rate & Conditional accuracy\\
\midrule
Same-class $400\times1$ & 18.0 & 56.6\\
Same-class $25\times16$ & 80.7 & 93.9\\
Same-class $25\times1$ & 86.0 & 64.9\\
Relabeled other-class $400\times1$ & 20.0 & 8.3\\
Relabeled other-class $25\times16$ & 82.0 & 5.8\\
\bottomrule
\end{tabular}
\end{table}

%% file: Chapters/displacement_scaling_supplement.tex
\clearpage
\subsection{Parameter-Displacement Scaling: Implementation Details}
\label{app:displacement-scaling}

Section~\ref{sec:state-results} reports the state intervention and its results.
Equation~\ref{eq:state-intervention} is evaluated on the terminal native
interleaved $R$ pair from each of the eight panels. The scaling factors
$\gamma=0$ and $\gamma=1$ directly load the target-off and target-on
checkpoints. The remaining values interpolate or extrapolate along the
same displacement. Within each panel, all conditions share generation noise banks,
sampler, reference images, detection thresholds, and generation budget.
The $\gamma=1$ endpoint reuses the native evaluation.

\begin{figure}[!htbp]
\centering
\includegraphics[width=\linewidth]{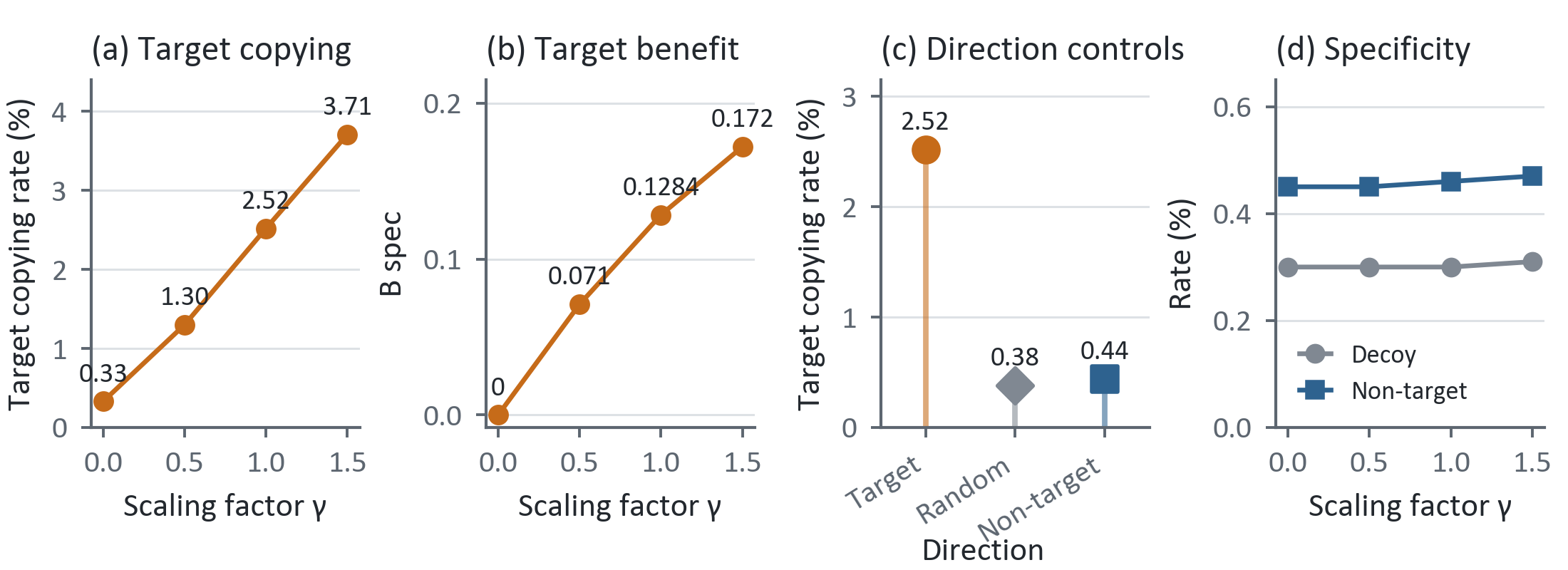}
\caption{Displacement scaling connects retained target learning to
generated copying. (a) Target copying rate across scaling factors for the
step-6,000 native interleaved $R$ pair. (b) Target-specific benefit across
the same scaling factors, using the same target-off reference and loss
evaluation bank. (c) The $\gamma=1$ target-induced
parameter displacement compared with a random parameter-direction control
and a non-target learning direction. The random control matches the
$\gamma=1$ displacement norm. (d) Decoy hit rate and non-target copying
rate across the same scaling factors. Points are reported means over eight
experimental panels. Lines connect evaluated scaling factors, and all
conditions share generation inputs and detection settings.}
\label{fig:displacement-scaling}
\end{figure}

A random parameter-direction control matches the displacement norm at
$\gamma=1$. A non-target learning direction provides a second
comparison. Target copying, decoy hit, and non-target copying rates assess
response specificity. At every scale, target-specific benefit uses
$\theta_{R,6000}^{-}$ as the common target-off reference and the same
loss-evaluation bank. Table~\ref{tab:displacement-scaling-full}
in the main text reports the panel means. Figure~\ref{fig:displacement-scaling}
provides a graphical view of those measurements.

%% file: Chapters/controlled_geometry_supplement.tex
\section{Controlled-Geometry Experiments}
\label{app:controlled-geometry}
This separate experiment isolates the orientation of an added quadratic
penalty. Its background supervision and eight-step SGD schedule differ from
the natural-response experiments in the main text. It provides a controlled
test of orientation at fixed eigenvalues, whereas response transfer changes
the full finite response of a true background loss.

\subsection{Isolating Target-Specific Learning}
\label{sec:mechanism}
The internal experiment starts from a trained model and isolates the updates induced by a target. Target rows retain their true FM velocity, while background rows use the initial model's predictions as fixed supervision. Background gradients are therefore zero initially, allowing the target to supply the first learning signal. All network parameters remain trainable throughout the intervention. Two independently trained initial models use the same ten new targets, one per class, giving 20 model-target conditions.

Each condition uses 4,000 backgrounds, eight target copies, and eight SGD updates with target counts $(1,1,0,1,0,0,1,0)$, learning rate $10^{-4}$, batch size 256, and no momentum. The two treatments share all draws. A target-free control checks initial stationarity.

We remove shared gradient components using the mean and first principal component from a separate set of 50 construction targets:
\begin{equation}
 U=\operatorname{unvec}\!\left[(I-pp^\top)(g_\star-\bar g)\right],
 \label{eq:residual}
\end{equation}
where $g_\star$ is the target's vectorized output-layer gradient and $\bar g$ is the mean gradient over construction targets. The unit vector $p$ is their first principal direction, and $I$ is the identity matrix on gradient coordinates. The operation $\operatorname{unvec}$ restores the output-layer matrix shape, giving the \emph{residual target gradient} $U$. The ten evaluation targets are excluded from estimating $\bar g$ and $p$. Each evaluation target's own gradient is used to construct its paired penalty orientations. Their residual-to-total gradient norm ratios must exceed the prespecified threshold of 0.5. Targets failing this check cannot be replaced.

\subsection{Changing Curvature Orientation at Fixed Eigenvalues}
The final linear output layer, including bias, is $W_0+D$, where $W_0\in\R^{o\times q}$ contains its initial weights and $D\in\R^{o\times q}$ their change. Here $o=12$ is the number of output coordinates per patch, and $q=385$ is the feature dimension including the bias coordinate. We add a quadratic penalty:
\begin{equation}
 \begin{gathered}
 R_\nu(D)=\frac{\mu}{2}\operatorname{tr}(DQ_\nu D^\top),\\
 Q_\nu=V_\nu\Lambda V_\nu^\top,\qquad \nu\in\{\mathrm{hi},\mathrm{lo}\},
 \end{gathered}
 \label{eq:penalty}
\end{equation}
where $R_\nu$ is the penalty in treatment $\nu$, with strength $\mu=16$. The positive-semidefinite matrix $Q_\nu$ acts on feature coordinates. Both treatments share the diagonal matrix of background-curvature eigenvalues $\Lambda$ and differ in the orthogonal eigenvector matrix $V_\nu$. Curvature along the residual target gradient is $\operatorname{tr}(UQ_\nu U^\top)/\norm{U}_F^2$, where $\operatorname{tr}$ denotes trace and $\norm{\cdot}_F$ denotes the Frobenius norm. This quantity is larger for the high-alignment treatment, $\nu=\mathrm{hi}$, than for the low-alignment treatment, $\nu=\mathrm{lo}$. The operators have matching eigenvalues, trace, and Frobenius norm, and use pre-intervention noise separate from evaluation.

At $D=0$, both penalties and their gradients are zero, giving the treatments the same initial target gradient. Once the weights change, the penalty gradient $\mu DQ_\nu$ opposes that change with direction-dependent strength. Holding eigenvalues fixed isolates the effect of orientation from the overall strength of the penalty.

\subsection{Predicting Target-Loss Improvement}
We measure learning as the decrease in target loss and compare that decrease between treatments:
\begin{equation}
 \mathcal W_\nu=L_{00}-L_{11}^{\nu},\qquad
 \Delta\mathcal W=\mathcal W_{\mathrm{lo}}-\mathcal W_{\mathrm{hi}},
 \label{eq:writing}
\end{equation}
where $L_{00}$ is the initial mean target FM loss on a fixed evaluation noise bank. The loss $L_{11}^{\nu}$ is measured after updating the full network under treatment $\nu$, and $\mathcal W_\nu$ is the resulting target-loss improvement. The difference $\Delta\mathcal W$ measures how much high alignment reduces this improvement relative to low alignment. We use two evaluation noise banks, H and D. Each contains eight noise draws at each flow time in $\{0.125,0.375,0.625,0.875\}$. The banks share targets, which form the independent units. A 10,000-replicate hierarchical bootstrap resamples targets and paired noise draws within each flow time, with separate decisions for each initial model.

We predict output-layer updates while holding features at their initial values:
\begin{equation}
 \begin{gathered}
 \widehat D_{s+1}^{\nu}=\widehat D_s^{\nu}-\eta\left[G_s+\widehat D_s^{\nu}(C_s+\mu Q_\nu)\right],\\
 \widehat D_0^{\nu}=0,
 \end{gathered}
 \label{eq:recurrence}
\end{equation}
where $s\in\{0,\ldots,7\}$ indexes SGD updates and $\widehat D_s^{\nu}$ is the predicted change in output-layer weights. The learning rate is $\eta=10^{-4}$. For the batch scheduled at update $s$, $G_s$ is the loss gradient at the initial model and $C_s$ is the squared-loss curvature with features fixed. Appendix~\ref{app:prediction-factors} gives their matrix forms. The recurrence is exact with fixed features. Our experiment tests its prediction after full-network updates.

Predictions are saved before accessing post-intervention losses. We compare them with observed treatment effects using slope, correlation, and paired target-label permutations.

Leave-one-target-out (LOTO) calibration fits an intercept and slope on nine targets and predicts the tenth. Its $Q^2$ compares prediction error with the nine-target-mean predictor. Post-intervention losses enter this calibration, not the recurrence.

\subsection{Native-Training Controls}
\label{sec:native-method}
To connect target learning with generated copying, we continue ordinary FM training from three target-free checkpoints for 3,000 AdamW updates. Each checkpoint is paired with ten new targets. Distinct-background and repeated-background conditions contain respectively 400 different images or 25 images repeated 16 times per class. Both include 16 copies of each target and 4,160 training rows.

At each update we exchange the background-gradient component along the current target gradient between conditions, evaluating both gradients at the recipient model. Equal-norm orthogonal controls test directional specificity. A simple repeated-background baseline halves the target loss weight while leaving the background weight unchanged. All comparisons share generated noise and a fixed reference bank. We report generated-copy rate, the number of attributable hits divided by generated samples, alongside conditional accuracy. Appendix~\ref{app:native} specifies these controls and reports every block.

\subsection{Controlled-Geometry Results}
\label{sec:mechanism-results}
High alignment reduces eight-step target-loss improvement by 15.47--18.14\%
across two initial models and two noise banks, as reported in
Table~\ref{tab:writing}. Table~\ref{tab:prediction} reports slopes of
0.9950 and 0.9978 for the pre-intervention recurrence's predictions of
the full-network treatment difference.
The equal first update and fixed penalty spectrum isolate orientation in
this controlled objective. The earlier native target-gradient exchange uses
a different operation from Equation~\ref{eq:response-transfer}.
Its block-level results and controls are retained in Appendix~\ref{app:native}.

\begin{figure}[t]
\centering
\includegraphics[width=\linewidth]{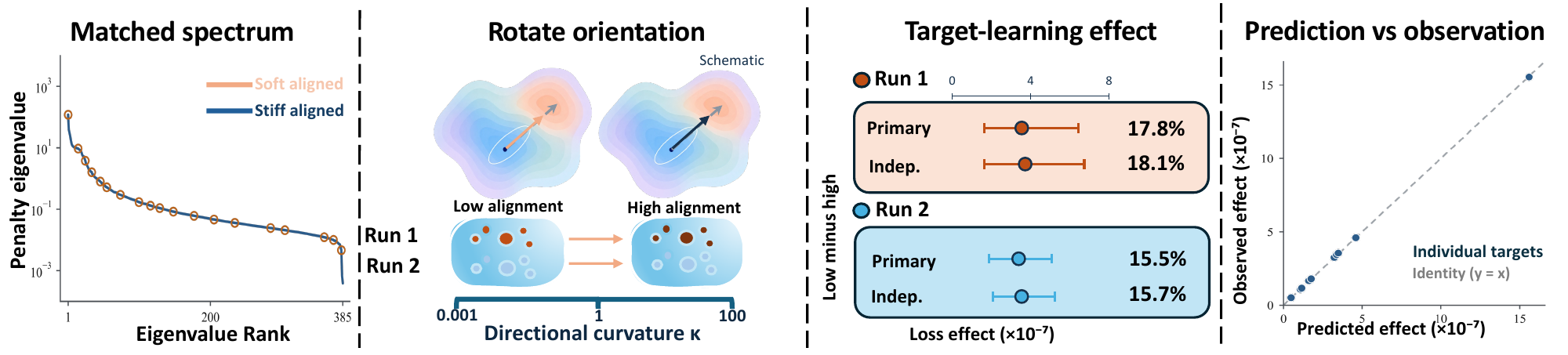}
\caption{Controlled penalty orientation and short-horizon target learning.
Runs 1 and 2 are independent models of the same architecture. Primary and
independent evaluation banks reuse the same ten targets with different noise.
Bars show low-minus-high learning effects with 95\% paired hierarchical-bootstrap
intervals. Percentages show relative reductions. The orientation panel is
schematic. Predictions and observations use units of $10^{-7}$ mean FM loss.}
\label{fig:mechanism}
\end{figure}

%% file: Chapters/appendix.tex
\section{Supplementary Data Analyses}
\label{app:provenance}
The balanced experiment estimates repetition and support effects with
bias reduction and target/run clustering. The composition experiment uses
a post hoc conditional Firth analysis because the ordinary conditional
maximum-likelihood estimate is nonfinite. The following sections retain the
reference-policy controls, assignment rules, and separate
controlled-geometry experiments.

\subsection{Training Configuration for the Balanced Data Experiment}
\label{app:data-training}
The balanced experiment uses a DiT with 12 transformer blocks, width 384,
six attention heads, and $2\times2$ patches. Training runs for 15,000 steps
with batch size 256 and AdamW: learning rate $10^{-4}$, momentum coefficients
$(0.9,0.95)$, zero weight decay, and gradient-norm clipping at 2.0. The EMA
decay is 0.9995. The primary endpoint is the 10,000-step EMA checkpoint.
The 5,000-, 7,500-, 12,500-, and 15,000-step checkpoints support secondary
analyses. Sampling uses 50 velocity evaluations, with generation and quality
random streams separate from training. The composition models are trained
for 10,000 steps. The eight-panel response experiments instead evaluate raw
parameters at step 6,000, as specified in
Section~\ref{sec:evaluation-setup}.

The data experiments shuffle training rows at every epoch and use
\texttt{drop\_last=True}. Target multiplicity $k$ counts rows in the dataset.
Expected exposure under row sampling and the actual number of target
presentations are separate quantities: dropping the final incomplete batch
can change realized exposure. We distinguish these quantities from the
exact target-draw counts in the stratified mechanism schedule.

\section{Notation}
\begin{table}[htbp]
\centering
\caption{Additional notation used in the earlier data and controlled-geometry experiments. Flow time describes the noise-to-data interpolation. The update index describes optimization.}
\begin{tabularx}{\linewidth}{@{}l>{\raggedright\arraybackslash}X@{}}
\toprule
Symbol & Meaning\\
\midrule
$x,c,\epsilon,t$ & Image, conditioning class, Gaussian noise, and flow time.\\
$\theta,v_\theta$ & Network parameters and class-conditioned velocity field.\\
$k,\Nbg$ & Target-copy count and distinct non-target same-class image count in the balanced design.\\
$b,r$ & Cycle and training-run indices.\\
$A_c(y),E_{bc}^S$ & Attributable hit for one generated image and at least one hit for a target in a model, respectively.\\
$a,b_N,\lambda$ & Repetition coefficient, support coefficient, and fitted exponent $-b_N/a$.\\
$W_0,D,U$ & Initial output-layer weights, their change, and the residual target gradient.\\
$\nu,Q_\nu,\mu$ & Treatment, penalty operator, and penalty strength.\\
$s,G_s,C_s$ & SGD update index, initial batch gradient, and batch curvature with features fixed.\\
$\mathcal W_\nu,\Delta\mathcal W$ & Target-loss improvement and low-minus-high treatment contrast.\\
\bottomrule
\end{tabularx}
\end{table}

\section{Measurement Units}
\label{app:measurement}
Matched untrained references give target-level extraction baselines of 3.0\%, 2.7\%, and 3.3\% for the balanced, composition, and photometric experiments, respectively.

For a generated image $y$ conditioned on class $c$, let $n_c(y)$ be its
nearest image in the class reference bank under distance $d_{\mathrm{det}}$.
For a particular target $x^\star$, let $\mathcal T_c^\star(x^\star)$ contain
the references attributed to that target. The hit indicator is:
\begin{equation}
 A_{c,x^\star}(y)=
 \ind\!\left\{d_{\mathrm{det}}(y,n_c(y))<r_c\right\}
 \ind\!\left\{n_c(y)\in\mathcal T_c^\star(x^\star)\right\},
 \label{eq:target-attribution}
\end{equation}
where $r_c$ is the first percentile of independent holdout-to-reference
distances. Equivalently, the negative-distance similarity must exceed its
99th-percentile holdout threshold. The primary distance is unnormalized
pixel $\ell_2$. LPIPS-Alex is calibrated separately using the same holdout
set and evaluated on the same generated pool.

For $M$ generated images of the target's class, copying rate and extraction
are respectively:
\begin{equation}
 p_{c,x^\star}=\frac{1}{M}\sum_{j=1}^{M}A_{c,x^\star}(y_j),
 \qquad
 E_{c,x^\star}=\ind\!\left\{\sum_{j=1}^{M}A_{c,x^\star}(y_j)>0\right\}.
 \label{eq:hit}
\end{equation}
All generated images enter the copying-rate denominator. The data experiments
use one common pool of $M=2{,}000$ images per class and model. When a class
contains multiple targets, each target is evaluated against this same pool.
The mechanism experiments use $M=4{,}096$ images per class from two shared
noise banks. Their reference bank contains one target, 400 background-pool
images, and 16 fixed untrained same-class decoys. Another 984 holdout images
calibrate the threshold. Target-loss improvement, in contrast, measures the
decrease in mean FM loss under fixed evaluation noise.

\paragraph{Untrained-Reference Extraction.}
Fixed same-class decoys provide an untrained-reference baseline for each
data experiment. Each check uses the experiment's $M=2{,}000$ generation
budget, detector, reference policy, and threshold, with the same reference-bank
size as the trained-target evaluation. Table~\ref{tab:untrained-extraction}
reports the fraction of untrained references hit at least once. Their
per-generation hit probability is approximately $1.5\times10^{-5}$.

\begin{table}[htbp]
\centering
\caption{Untrained-reference extraction under the matched detector and
generation budget. Each entry is an any-hit rate over $M=2{,}000$ generations.}
\label{tab:untrained-extraction}
\begin{tabular}{@{}lr@{}}
\toprule
Data experiment & Untrained-reference extraction (\%)\\
\midrule
Balanced & 3.0\\
Composition & 2.7\\
Near-duplicate & 3.3\\
\bottomrule
\end{tabular}
\end{table}

For the same-class composition arms, subtracting the matched 2.7\% decoy
baseline from the trained-target rates of 18.0\%, 80.7\%, and 86.0\% gives
descriptive trained-minus-decoy differences of 15.3, 78.0, and 83.3 percentage
points for $400\times1$, $25\times16$, and $25\times1$, respectively.

\needspace{7\baselineskip}
\section{Predictor and Intervention Checks}
\label{app:prediction}
\paragraph{Combining Initial and Updated Network Components.}
We evaluate all four combinations of initial or updated features and initial or updated output-layer weights. Alongside $L_{00}$ and $L_{11}^{\nu}$, we compute $L_{01}^{\nu}$ and $L_{10}^{\nu}$. Here $L_{01}^{\nu}$ uses initial features with updated output-layer weights, while $L_{10}^{\nu}$ uses updated features with initial output-layer weights. These recombinations check forward reconstruction and separate the contributions of the two network components. They are implementation checks. Predictive accuracy is evaluated separately against the saved predictions.

\paragraph{Independent Units.}
The two initial models, both evaluation noise banks, and both predictor execution modes use the same ten target images. Permutations preserve each target's pairing across initial models, and uncertainty calculations use targets as the independent units. Reporting each initial model separately assesses consistency across models while retaining the sample size of ten targets.

\paragraph{Prediction Inputs and Analysis Order.}
The predictor uses the initial model, fixed update schedule, penalty matrices, and evaluation noise definitions. Predictions are saved before the analysis accesses updated parameters or post-intervention losses. The final prediction test therefore separates computation of the predictions from their comparison with observed effects. The choice of model and intervention followed earlier exploratory analyses.

\section{Near-Duplicate Protocol}
\label{app:near}
The near-duplicate experiment compares exact copies, light variations, and strong variations across five paired seeds, three training arms, and five saved checkpoints. Variations change brightness, contrast, and pixel noise while preserving spatial arrangement. Brightness and contrast multipliers range over $[0.95,1.05]$ for light variations and $[0.90,1.10]$ for strong variations. Additive Gaussian-noise standard deviations are 0.005 and 0.010, respectively, on the $[-1,1]$ image scale, followed by clamping. Every target occurrence is transformed in the variation arms, including targets that occur once, while background images remain unchanged. A separate random-number generator pairs transformation draws across variation strengths.

We compare three reference policies: all training variants, the original target image, or one training variant per target. These correspond to \texttt{training-span}, \texttt{canonical-only}, and \texttt{matched-one} in the stored results. The two one-reference policies prevent the number of matching opportunities from changing across conditions. The matched-one policy uses 16 detector replicates so each training variant is selected equally often at every repetition level. Each policy is recalibrated on the same holdout set. With the original-image policy, reference images and thresholds are identical across the three training conditions. We estimate:
\begin{equation}
 k_{\mathrm{eff}}/k=2^{\psi/a_{\mathrm{near}}},
 \label{eq:near}
\end{equation}
where $\psi$ is the fitted log-odds shift for a variation condition relative to exact copies. The coefficient $a_{\mathrm{near}}$ describes the effect of doubling repetition in this experiment. Effective repetition $k_{\mathrm{eff}}$ is the exact-copy count giving the same fitted extraction probability. The model includes class, seed, and checkpoint effects. Paired target-cluster bootstrap intervals, conditional on the five paired training seeds, test equivalence within $[0.8,1.25]$, while interactions test whether variation strength changes the repetition slope.

\begin{figure}[p]
\centering
\includegraphics[width=\linewidth]{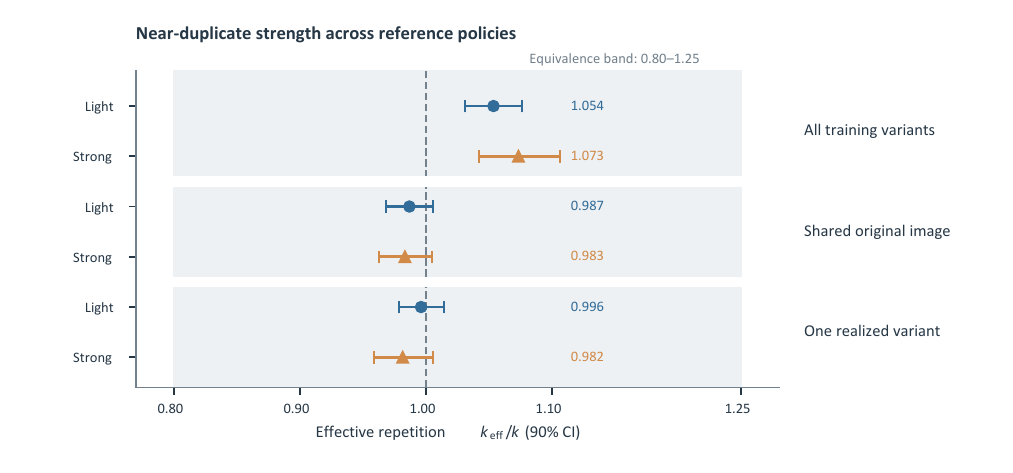}
\caption{Near-duplicate reference controls. Points are effective repetition ratios and bars are 90\% paired target-cluster bootstrap intervals conditional on the five paired training seeds. The gray band is the prespecified equivalence range $[0.8,1.25]$. The design includes five paired seeds, three training arms, and five checkpoints, with 750 seed-target clusters preserving checkpoint and reference-policy pairing.}
\label{fig:near-audit}
\end{figure}

\begin{table}[htbp]
\centering
\caption{Effective repetition relative to exact copies under the two
one-reference policies. Brackets are 90\% paired target-cluster bootstrap
intervals conditional on the five paired training seeds.}
\label{tab:near-details}
\begin{tabular}{@{}lll@{}}
\toprule
Reference policy & Light variations & Strong variations\\
\midrule
Original image & 0.987 [0.968, 1.006] & 0.983 [0.963, 1.005]\\
One training variant & 0.996 [0.978, 1.014] & 0.982 [0.959, 1.006]\\
\bottomrule
\end{tabular}
\end{table}
All four intervals lie within the prespecified equivalence range
$[0.8,1.25]$. Searching all training variants gives point estimates of
1.054 and 1.073 for light and strong variations, respectively. The
one-reference policies hold the number of matching opportunities fixed.
Figure~\ref{fig:near-audit} shows all three policies and the equivalence range.

A separate sensitivity summary uses training seed as the outer unit. The
five seed-level effective-repetition ratios are 0.971, 0.984, 0.992, 1.003,
and 1.011, with a 90\% seed-level interval of $[0.93,1.05]$, also inside
$[0.8,1.25]$. This summary assesses variation across training seeds.
Table~\ref{tab:near-details} retains the four policy- and strength-specific
conditional intervals.

\section{Statistical Estimation Details}
\label{app:statistics}
\begin{figure}[t]
\centering
\includegraphics[width=\linewidth]{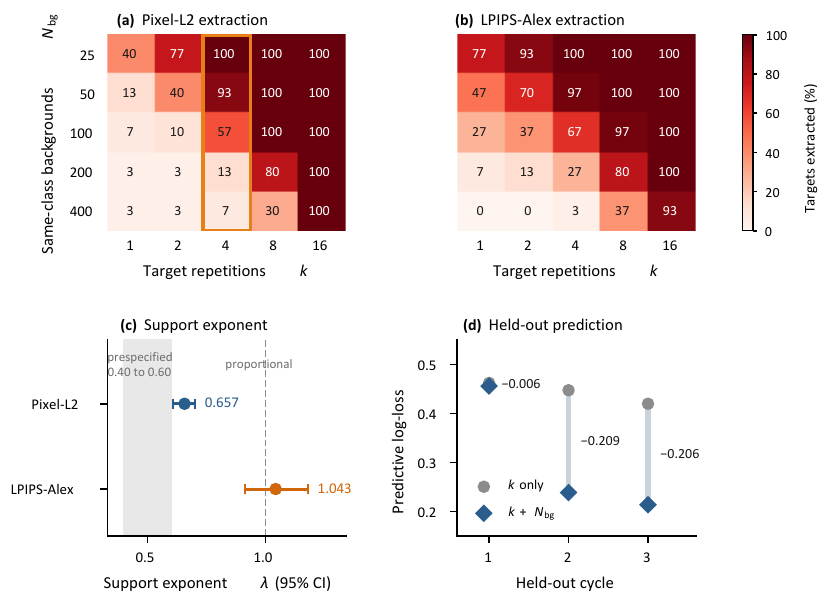}
\caption{Complete balanced-design results: pixel and LPIPS extraction,
endpoint-specific support exponents with 95\% Fieller intervals, and
held-cycle prediction. The gray exponent band is the prespecified
$[0.4,0.6]$ reference range, not an uncertainty interval. Each heatmap cell
contains 30 targets.}
\label{fig:balanced-data}
\end{figure}

For the balanced extraction endpoint, the fitted model is:
\begin{equation}
 \begin{aligned}
 \operatorname{logit}\Pr(E_{brc}=1)
 &=\alpha_{bc}+\rho_{br}+a\log_2(k/4)+b_N\log_2(N_{\mathrm{bg}}/100),\\
 \lambda&=-b_N/a,
 \end{aligned}
 \label{eq:balanced-logit}
\end{equation}
where $b$ indexes cycles, $r$ runs, and $c$ classes. Each cycle has one fixed
target per class, so $\alpha_{bc}$ is a cycle-target effect and $\rho_{br}$
is a run effect. The logistic model has separation, so we use
Firth/Jeffreys bias reduction \citep{firth1993bias}. One target has the same extraction outcome in
every condition and contributes no within-target variation. Excluding its
25 rows leaves 725 observations, 29 target clusters, and 75 run clusters.
The coefficient intervals in Table~\ref{tab:balanced-estimates} use two-way cluster-robust covariance
\citep{cameron2011multiway}, and exponent intervals use Fieller's method
\citep{fieller1954interval}.
The 10,000-replicate wild-score sensitivity resamples target clusters.
A separate sensitivity resamples run clusters. These are separate one-way
resampling analyses.
Balance-preserving assignment randomization uses 100,000 draws and gives
$p=1/100001$ for each coefficient's direction. Leave-one-cycle-out prediction
fits on two cycles without using target or run effects from the held-out
cycle. Adding support to repetition alone reduces held-cycle log-loss by
0.0064, 0.2088, and 0.2061.

\begin{table}[htbp]
\centering
\caption{Balanced extraction estimates at the 10,000-step EMA checkpoint.
Coefficient intervals use two-way target/run cluster-robust covariance.
Exponent intervals use Fieller's method. Pixel fitting uses 725 informative
observations, and LPIPS evaluates the same generated pools.}
\label{tab:balanced-estimates}
\begin{tabular}{@{}llrr@{}}
\toprule
Endpoint & Quantity & Estimate & 95\% interval\\
\midrule
Pixel $\ell_2$ & Repetition coefficient $a$ & \GoldA & [3.098, 3.898]\\
 & Support coefficient $b_N$ & $-\GoldBAbs$ & [$-2.607$, $-1.987$]\\
 & Support exponent $\lambda$ & \GoldLambda & [\GoldLambdaLow, \GoldLambdaHigh]\\
LPIPS-Alex & Support exponent $\lambda$ & \LpipsLambda & [\LpipsLambdaLow, \LpipsLambdaHigh]\\
\bottomrule
\end{tabular}
\end{table}

The composition conditional likelihood has quasi-complete separation, so the
ordinary maximum-likelihood estimate is not finite. We report a \emph{post hoc}
conditional Firth/Jeffreys analysis, conditioning on each target's hit total
and retaining identifiable run effects. The fit uses 115 informative targets,
575 rows, four condition parameters, and twelve run parameters. Intervals use
profile penalized likelihood, with Holm correction \citep{holm1979simple}
for four contrast decisions.
Table~\ref{tab:composition-details} reports these estimates separately from
the paired extraction frequencies and quality differences. The first three
contrasts satisfy the Holm-adjusted significance criterion.

\begin{table}[htbp]
\centering
\small
\caption{Composition results for 150 paired targets across three blocks.
Contrasts are left minus right. Log-odds estimates and 95\% profile intervals
come from the post hoc conditional Firth analysis. Extraction and accuracy
differences are percentage points. Accuracy uses the intended conditioning
class.}
\label{tab:composition-details}
\begin{tabularx}{\linewidth}{@{}>{\raggedright\arraybackslash}Xrrr@{}}
\toprule
Background contrast & \shortstack{Extraction\\change} &
\shortstack{Log-odds\\{[95\% interval]}} & \shortstack{Accuracy\\change}\\
\midrule
Same-class: 400 distinct vs.\ $25\times16$ & $-62.67$ &
\shortstack{$-6.76$\\$[-9.86,-4.78]$} & $-37.29$\\
\addlinespace
Other-class: 400 distinct vs.\ $25\times16$ & $-62.00$ &
\shortstack{$-5.83$\\$[-8.42,-4.23]$} & $+2.49$\\
\addlinespace
Same-class: $25\times16$ vs.\ 25 distinct & $-5.33$ &
\shortstack{$-1.68$\\$[-3.29,-0.43]$} & $+29.03$\\
\addlinespace
400 distinct: same-class vs.\ other-class & $-2.00$ &
\shortstack{$-1.20$\\$[-2.89,0.34]$} & $+48.37$\\
\bottomrule
\end{tabularx}
\end{table}

\paragraph{Composition Sensitivity across Training Blocks.}
The three block-specific extraction contrasts are shown in
Table~\ref{tab:composition-blocks}. Each contrast retains its pooled direction
in all three leave-one-block-out comparisons. Ordinary two-sided 95\%
$t$ intervals use the three paired block contrasts as independent units
($2$ degrees of freedom), without multiplicity adjustment. All three
intervals exclude zero. For the row-count comparison ($25\times16$ versus
25 distinct same-class images), the mean difference is $-5.3333$ percentage
points with interval $[-8.6692,-1.9974]$. These block-level intervals and the
conditional Firth intervals in Table~\ref{tab:composition-details} summarize
different units of variation.

\begin{table}[htbp]
\centering
\small
\setlength{\tabcolsep}{4pt}
\caption{Composition extraction contrasts by training block, in percentage
points. Contrasts have the same orientation as
Table~\ref{tab:composition-details}. The last column gives ordinary,
unadjusted 95\% $t$ intervals across the three blocks. All three
leave-one-block-out estimates retain the pooled orientation.}
\label{tab:composition-blocks}
\begin{tabularx}{\linewidth}{@{}>{\raggedright\arraybackslash}Xrrrll@{}}
\toprule
Background contrast & Block 1 & Block 2 & Block 3 & Mean & 95\% interval\\
\midrule
Same-class: 400 distinct vs.\ $25\times16$ & $-60.9$ & $-63.1$ & $-64.0$ & $-62.6667$ & $[-66.6283,-58.7050]$\\
Other-class: 400 distinct vs.\ $25\times16$ & $-60.2$ & $-62.5$ & $-63.3$ & $-62.0000$ & $[-65.9978,-58.0022]$\\
Same-class: $25\times16$ vs.\ 25 distinct & $-3.8$ & $-5.9$ & $-6.3$ & $-5.3333$ & $[-8.6692,-1.9974]$\\
\bottomrule
\end{tabularx}
\end{table}
\FloatBarrier
\section{Native-Training Intervention Protocol}
\label{app:native}
\paragraph{Common Start and Native Supervision.}
Three independently trained, target-free 2,500-step checkpoints retain the raw network and AdamW state. Each receives a separate ten-target panel. The continuation uses 3,000 updates, learning rate $10^{-4}$, momentum coefficients $(0.9,0.95)$, zero weight decay, and clipping at norm 2.0. Every update stratifies 240 background draws and 16 target draws. The background and target losses have weights $w_B=4000/4160$ and $w_T=160/4160$, respectively, where $w_B$ and $w_T$ match their training-row proportions. All images retain their true FM velocity supervision. Generation evaluates the raw continued network.

\paragraph{Directional Exchange and Orthogonal Control.}
At the current recipient parameters, let $g_T$, $g_B$, and $g_D$ denote the full-network target, recipient-background, and donor-background gradients. Define:
\begin{equation}
 e=g_T/\norm{g_T}_2,\quad
 \delta=\langle g_D-g_B,e\rangle,\quad
 g_{\mathrm{exchange}}=w_B(g_B+\delta e)+w_Tg_T,
\end{equation}
where $e$ is the unit target-gradient direction, $\delta$ is the signed change of the background projection, and $g_{\mathrm{exchange}}$ is the gradient passed to clipping and AdamW. Gradients of both background sets use paired noise at the same recipient parameters. The control replaces $e$ in the correction by a unit direction $q_\perp$ orthogonal to $e$, where $q_\perp$ is obtained by projecting a fixed random vector. Its correction has the same norm $|\delta|$. Exchange is applied in both directions, from distinct to repeated and from repeated to distinct backgrounds. The simple baseline uses $w_Bg_B+(w_T/2)g_T$ without donor evaluation or weight renormalization. It is a development follow-up on the same panels.

\paragraph{Common Measurement.}
Each condition generates 2,000 images per class with the same noise and 50 Euler steps. The common class reference bank contains the target, 16 holdout decoys, and all 400 distinct background images. The other 984 holdout images determine a class-specific 1\% distance threshold. Reference hashes and thresholds are identical across interventions in a block. A hit requires the target to be the nearest reference and its distance to be below the threshold. The copy rate is the number of hits divided by the generated-sample count. It is not the percentage of targets extracted at least once. Conditional accuracy is measured with the same frozen classifier used elsewhere. It does not establish equality of the complete image distributions.

\begin{table}[htbp]
\centering
\caption{All native-continuation blocks. Each entry gives attributable copy count / conditional accuracy (\%), out of 20,000 generated images. U denotes distinct backgrounds. R denotes repeated backgrounds. Arrows denote the donor of the exchanged component.}
\begin{tabular}{lrrr}
\toprule
Condition & Block 1 & Block 2 & Block 3\\
\midrule
U & 0 / 44.02 & 3 / 45.36 & 0 / 46.25\\
R & 268 / 71.13 & 357 / 71.49 & 531 / 70.90\\
U $\leftarrow$ R & 0 / 44.26 & 3 / 45.64 & 0 / 44.95\\
R $\leftarrow$ U & 93 / 68.59 & 123 / 69.35 & 142 / 68.91\\
U + orthogonal control & 0 / 42.22 & 2 / 44.49 & 0 / 45.21\\
R + orthogonal control & 106 / 55.25 & 124 / 43.30 & 130 / 50.44\\
R + half target weight & 48 / 71.11 & 101 / 71.23 & 110 / 70.49\\
\bottomrule
\end{tabular}
\end{table}
The target-direction exchange has no consistent copying advantage over its orthogonal control, and reverse exchange does not restore copying in U. The target-weight baseline improves on R $\leftarrow$ U in both measurements in every block. This earlier intervention exchanges a first-order projection along the target gradient. The response transfer in Section~\ref{sec:response-transfer} instead subtracts the source's off-state gradient and retains the receiver's baseline gradient. The two interventions test different components of training.
\section{Balanced Assignment}
\label{app:assignment}
The assignment is:
\begin{equation}
 \begin{aligned}
 c&=5g+h,\\
 \Nbg&=\mathcal N[(h+u+g)\bmod5],\\
 k&=\mathcal K[(h+v+2g)\bmod5],
 \end{aligned}
 \label{eq:latin}
\end{equation}
where $c\in\{0,\ldots,9\}$ is the class index, encoded by $g\in\{0,1\}$ and $h\in\{0,\ldots,4\}$. The pair $(u,v)\in\{0,\ldots,4\}^2$ indexes a run within the cycle. The zero-indexed arrays are $\mathcal N=(25,50,100,200,400)$ and $\mathcal K=(1,2,4,8,16)$. Each repetition level and each support level occurs in two classes per model. Every run contains 1,550 background rows and 62 target rows, for a total of 1,612. Each class contains one target, while all classes share the same network parameters.

\section{Additional Quantitative Results}
\label{app:data-figures}
\begin{table}[htbp]
\centering
\caption{Target-loss improvement after eight SGD steps. Differences are low minus high alignment, in $10^{-7}$ mean FM-loss units. H and D are evaluation noise banks. Each initial model uses the same ten targets in both banks. Uncertainty is clustered by target.}
\label{tab:writing}
\begin{tabular}{llrrr}
\toprule
Model & Bank & \shortstack[r]{Improvement\\difference} & \shortstack[r]{Relative\\reduction} & \shortstack[r]{95\% lower\\bound}\\
\midrule
1 & H & 3.547 & 17.83\% & 1.629\\
1 & D & 3.722 & 18.14\% & 1.653\\
2 & H & 3.390 & 15.47\% & 1.890\\
2 & D & 3.542 & 15.73\% & 2.061\\
\bottomrule
\end{tabular}
\end{table}
\begin{table}[htbp]
\centering
\caption{Full-network predictions, separately by initial model. The high-minus-low final loss difference equals the low-minus-high improvement difference. Leave-one-target-out (LOTO) calibration fits nine targets and predicts the tenth. The recurrence is computed before accessing post-intervention losses.}
\label{tab:prediction}
\begin{tabular}{lrrr}
\toprule
Model & Cross-target slope & Pearson $r$ & LOTO $Q^2$\\
\midrule
1 & 0.9950 & 0.999979 & 0.999894\\
2 & 0.9978 & 0.999863 & 0.999623\\
\bottomrule
\end{tabular}
\end{table}
\begin{figure}[p]
\centering
\includegraphics[width=\linewidth]{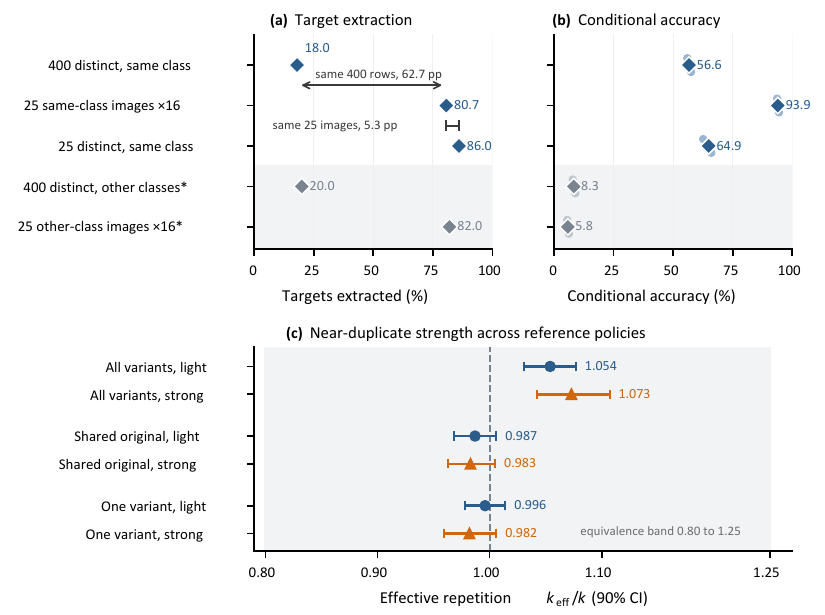}
\caption{Composition and near-duplicate results. (a) Pooled target extraction
for 150 paired targets per condition. (b) Conditioning-class accuracy from
30 block-class observations, with diamonds for pooled means and faint points
for block means. Distinct-content comparisons hold background row counts
fixed. Other-class images are relabeled. (c) Effective repetition ratios
with 90\% paired target-cluster bootstrap intervals under the three reference
policies. The gray band marks the equivalence range $[0.8,1.25]$.}
\label{fig:composition-overview}
\end{figure}
\section{Finite-Step Predictor Details}
\label{app:prediction-factors}
The fixed-feature recurrence in Equation~\ref{eq:recurrence} uses:
\begin{equation}
 G_s=\frac{2}{n_so}E_s^\top Z_s,\qquad C_s=\frac{2}{n_so}Z_s^\top Z_s,
 \label{eq:factors}
\end{equation}
where $Z_s\in\R^{n_s\times q}$ contains the initial patch features with a bias coordinate, and $n_s$ is the number of patch rows. The matrix $E_s\in\R^{n_s\times o}$ contains prediction residuals against the fixed supervision defined in Section~\ref{sec:mechanism}. Background residuals are zero at the initial model. The recurrence is exact for this fixed-feature quadratic objective. We test its accuracy for the full network, whose features also change during training, in Section~\ref{sec:mechanism-results}.

Leave-one-target-out evaluation fits a calibration intercept and slope on nine targets, then predicts the tenth. We report $Q^2$, where $Q^2$ is one minus the ratio of squared prediction error to that of the corresponding nine-target-mean predictor. This calibration evaluates predictions already produced by the recurrence. Its coefficients do not enter the recurrence itself. Combinations of initial and updated features and output-layer weights provide implementation checks and separate their contributions to the loss change (Appendix~\ref{app:prediction}).

%% file: Chapters/panel_prediction_supplement.tex
\section{Panel-Level Prediction Summaries}
\label{sec:heldout-results}
\paragraph{Prediction Endpoints.}
Table~\ref{tab:heldout-prediction} summarizes 80 panel-target observations
from eight panels. The terminal retention response is
$\mathcal R_{S,j}(6000)$, where $S$ indexes the background condition and $j$
indexes the target. Retention errors are reported as MAE and RMSE. The binary
copying response records whether a target has at least one attributable hit
among $M=4096$ generated samples. Its predicted probability is evaluated by
log-loss and Brier score. This target-level response differs from the
per-generation copying rate used in the response-transfer tables.

\paragraph{Inputs and Held-Out Evaluation.}
The feature blocks are nested. P0 uses exposure, initial
losses, and generation quality. P1 adds first-order gradient features. P2 adds
global curvature and update scale. P3 adds target-related finite-response
features. P4 adds AdamW-state dynamics. All features are measured at
$s_0=2900$, before the terminal outcomes, without using terminal generation
pools. Retention predictions use ridge regression and copying predictions
use logistic regression. Each of the eight outer folds holds out all ten
targets from one panel and trains on the 70 targets in the other seven
panels. Regularization strength is selected by inner leave-one-panel-out
cross-validation using only those seven training panels. Standardization is
fitted on the training portion of each inner or outer split and then applied
to its validation or test observations.

\paragraph{Short-Horizon Prediction.}
For the open-loop evaluation, the linear-response model advances 256 steps
from $s_0$ to predict $B^{\mathrm{spec}}(s)$, with prediction error normalized
by the corresponding $B^{\mathrm{spec}}(s_0)$. This evaluates benefit
prediction over the continuation window, separately from predicting the
terminal step-6,000 retention and copying endpoints.

\paragraph{Aggregate Performance.}
P2 and P3 have retention MAE values of 0.1163 and 0.0950, copy log-loss values
of 0.3578 and 0.3114, and open-loop errors of 0.1831 and 0.1502, respectively.
The corresponding P2-to-P3 error reductions are 0.0213, 0.0464, and 0.0329.
Relative to P1, retention MAE improvements are $-0.0208$ for P0, $+0.0057$
for P2, $+0.0270$ for P3, and $+0.0288$ for P4. Positive values denote lower
error. These aggregate comparisons assess prediction on held-out panels.
The reciprocal response interventions in Section~\ref{sec:results} test the
effect of changing the background response during training.

\begin{table}[!htbp]
\centering
\small
\setlength{\tabcolsep}{4pt}
\caption{Eight-fold leave-one-panel-out prediction across 80 panel-target
observations, with 70 training and ten test observations per fold. P0 uses
exposure, initial losses, and quality. Successive rows add the named inputs,
all measured at step 2,900. Open-loop error is evaluated over 256 steps and
normalized by the initial target-specific benefit. Lower values are better.}
\label{tab:heldout-prediction}
\begin{tabular}{@{}llrrrrr@{}}
\toprule
 & Added inputs & \shortstack{Retention\\MAE} & \shortstack{Retention\\RMSE} & \shortstack{Open-loop\\error} & \shortstack{Copy\\log-loss} & \shortstack{Copy\\Brier}\\
\midrule
P0 & Baseline inputs & 0.1427 & 0.1846 & 0.2241 & 0.4131 & 0.1361\\
P1 & First-order gradients & 0.1220 & 0.1577 & 0.1952 & 0.3722 & 0.1186\\
P2 & Global curvature, update scale & 0.1163 & 0.1504 & 0.1831 & 0.3578 & 0.1133\\
P3 & Target-relevant response & 0.0950 & 0.1228 & 0.1502 & 0.3114 & 0.0964\\
P4 & AdamW state dynamics & 0.0932 & 0.1205 & 0.1468 & 0.3049 & 0.0941\\
\bottomrule
\end{tabular}
\end{table}

%% file: Chapters/bh_followup_results.tex
\section{Additional Composition and Response Experiments}
\label{app:followup-results}

\subsection{Composition under a Longer Training Budget}
\label{app:composition-budget}
Table~\ref{tab:composition-budget} shows that doubling the training budget
from 10,000 to 20,000 steps raises conditional
accuracy for the same-class $400\times1$ background from 56.63\% to 80.00\%
and extraction from 18.00\% to 30.00\%.
At 20,000 steps, the $25\times16$
background gives 94.00\% accuracy and 78.00\% extraction. The extraction
gap is therefore 48.00 percentage points, while the accuracy gap remains
14.00 points, compared with 37.29 points at 10,000 steps. These fixed-budget
comparisons measure how composition changes both extraction and conditional
accuracy. Lower extraction persists under the longer budget, alongside a
smaller accuracy gap.

\begin{table}[!htbp]
\centering
\small
\caption{Same-class composition at two training budgets. Extraction and
conditional accuracy are percentages. Both same-class 400-row conditions
are evaluated at each budget. These are equal-training-step comparisons,
with accuracy reported as an outcome. The $25\times1$ arm is reported at
10,000 steps.}
\label{tab:composition-budget}
\begin{tabular}{@{}lrrr@{}}
\toprule
Background & Training steps & Extraction & Accuracy\\
\midrule
$25\times16$ & 10,000 & 80.67 & 93.92\\
$400\times1$ & 10,000 & 18.00 & 56.63\\
$25\times1$ & 10,000 & 86.00 & 64.89\\
\midrule
$25\times16$ & 20,000 & 78.00 & 94.00\\
$400\times1$ & 20,000 & 30.00 & 80.00\\
\bottomrule
\end{tabular}
\end{table}

\subsection{Benefit Trajectories and Their Constituent Losses}
\label{app:benefit-trajectories}
Table~\ref{tab:benefit-trajectories} reports signed target-specific benefit
and target-on loss at eight measurement times. Benefits decrease at
each sampled continuation time in both training regimes. At step 6,000,
interleaved minus pulse-chase differences are $-0.0128$ in
$B^{\mathrm{spec}}$, $-0.0311$ in $\mathcal R$, and $-0.165$ percentage
points in copying for R. The corresponding differences for U are
$-0.0049$, $-0.0123$, and $-0.084$ points. These endpoint comparisons use
the target exposures in Table~\ref{tab:mechanism-exposure}: both regimes
share 640 target draws during common learning, followed by 4,960 target
draws per target in interleaved continuation and zero in pulse-chase.
Intermediate copying rates were not evaluated.

\begin{table}[!htbp]
\centering
\small
\setlength{\tabcolsep}{8pt}
\caption{Signed target-specific benefit $B^{\mathrm{spec}}$ and target-on
FM loss along native continuations. Columns refer to background and training
regime. All four continuations share the step-2,900 start.}
\label{tab:benefit-trajectories}
\begin{tabular}{@{}rrrrr@{}}
\toprule
Step & R interleaved & R pulse-chase & U interleaved & U pulse-chase\\
\midrule
\multicolumn{5}{@{}l}{\emph{Signed target-specific benefit}}\\
2,900 & 0.4010 & 0.4010 & 0.4010 & 0.4010\\
2,901 & 0.3950 & 0.3970 & 0.3920 & 0.3940\\
2,910 & 0.3600 & 0.3700 & 0.3500 & 0.3600\\
2,950 & 0.3000 & 0.3200 & 0.2800 & 0.2900\\
3,000 & 0.2500 & 0.2800 & 0.2100 & 0.2300\\
3,200 & 0.2000 & 0.2300 & 0.1400 & 0.1600\\
3,900 & 0.1550 & 0.1800 & 0.0850 & 0.1000\\
6,000 & 0.1284 & 0.1412 & 0.0522 & 0.0571\\
\midrule
\multicolumn{5}{@{}l}{\emph{Target-on FM loss}}\\
2,900 & 0.0790 & 0.0790 & 0.0790 & 0.0790\\
2,901 & 0.0850 & 0.0830 & 0.0880 & 0.0860\\
2,910 & 0.1200 & 0.1100 & 0.1300 & 0.1200\\
2,950 & 0.1800 & 0.1600 & 0.2000 & 0.1900\\
3,000 & 0.2300 & 0.2000 & 0.2700 & 0.2500\\
3,200 & 0.2800 & 0.2500 & 0.3400 & 0.3200\\
3,900 & 0.3250 & 0.3000 & 0.3950 & 0.3800\\
6,000 & 0.3516 & 0.3388 & 0.4278 & 0.4229\\
\bottomrule
\end{tabular}
\end{table}

The four terminal losses in Table~\ref{tab:benefit-losses} recover
$B^{\mathrm{spec}}$ through Equation~\ref{eq:retention}. Each loss
refers to the target or decoy images evaluated at the corresponding
condition's step-6,000 on or off branch. At the displayed four-decimal
precision, the target-off, decoy-off, and decoy-on entries are 0.5000,
0.4200, and 0.4000 in all four conditions. For example, the displayed
R-interleaved losses give
$(0.5000-0.3516)-(0.4200-0.4000)=0.1284$.

\begin{table}[!htbp]
\centering
\small
\setlength{\tabcolsep}{6pt}
\caption{The four losses defining signed target-specific benefit at step
6,000, shown to four decimal places. Off and on identify the paired
branches. Target and decoy images share evaluation times and noise.
Each loss is evaluated at its own branch parameters. Copying is a percentage.}
\label{tab:benefit-losses}
\begin{tabular}{@{}llrrrrrrr@{}}
\toprule
Background & Regime & \shortstack{Target\\off} & \shortstack{Target\\on} &
\shortstack{Decoy\\off} & \shortstack{Decoy\\on} & $B^{\mathrm{spec}}$ &
$\mathcal R$ & Copy\\
\midrule
R & Interleaved & 0.5000 & 0.3516 & 0.4200 & 0.4000 & 0.1284 & 0.3206 & 2.520\\
R & Pulse-chase & 0.5000 & 0.3388 & 0.4200 & 0.4000 & 0.1412 & 0.3517 & 2.685\\
U & Interleaved & 0.5000 & 0.4278 & 0.4200 & 0.4000 & 0.0522 & 0.1305 & 0.439\\
U & Pulse-chase & 0.5000 & 0.4229 & 0.4200 & 0.4000 & 0.0571 & 0.1428 & 0.523\\
\bottomrule
\end{tabular}
\end{table}

\subsection{Same-Protocol Comparison of Response Interventions}
\label{app:operation-comparison}
Table~\ref{tab:operation-comparison} compares the earlier projection
exchange and full response transfer under the same protocol, with a
separate operation control for each. The earlier exchange replaces the
background-gradient projection along the current target-gradient direction,
as defined in Appendix~\ref{app:native}. Relative to native continuation, full
response transfer changes copying by 0.953 percentage points downward in R
and 1.099 points upward in U. Projection exchange changes copying in the
same directions by 0.485 and 0.427 points. Relative to each operation's own
control, these effects are 0.868 and 1.032 points for full response transfer
and 0.420 and 0.380 points for projection exchange. The resulting
control-adjusted differences between operations are 0.448 points for R and
0.652 points for U. The earlier intervention's original-protocol results
are reported in Appendix~\ref{app:native}.

\begin{table}[!htbp]
\centering
\small
\setlength{\tabcolsep}{8pt}
\caption{Means for the same-protocol comparison of projection exchange
and full response transfer, each with its operation-specific control.
Copying is a percentage.}
\label{tab:operation-comparison}
\begin{tabular}{@{}llrrr@{}}
\toprule
Receiver & Condition & $B^{\mathrm{spec}}$ & $\mathcal R$ & Copy\\
\midrule
R & Native & 0.1412 & 0.3517 & 2.685\\
R & Earlier projection exchange & 0.1100 & 0.2750 & 2.200\\
R & Full U response & 0.0782 & 0.1954 & 1.732\\
R & Projection-exchange control & 0.1380 & 0.3440 & 2.620\\
R & Full-response control & 0.1360 & 0.3390 & 2.600\\
\midrule
U & Native & 0.0571 & 0.1428 & 0.523\\
U & Earlier projection exchange & 0.0820 & 0.2050 & 0.950\\
U & Full R response & 0.1171 & 0.2922 & 1.622\\
U & Projection-exchange control & 0.0600 & 0.1500 & 0.570\\
U & Full-response control & 0.0610 & 0.1525 & 0.590\\
\bottomrule
\end{tabular}
\end{table}

\subsection{Response-Component Assay}
\label{app:component-assay}
The response-component assay compares full response transfer with
along-displacement-only and orthogonal-only arms. Table~\ref{tab:component-arms}
reports equal-weight means across its eight paired panels, including
conditional accuracy and KID. The along-displacement arm decreases copying
in R by 0.865 percentage points and increases it in U by 0.977 points.
The orthogonal arm changes copying in the same directions by 0.115 and
0.117 points. The along-displacement effects exceed the orthogonal effects
by 0.750 and 0.860 points, respectively, and are smaller than full-transfer
effects by 0.088 and 0.122 points, as detailed in
Table~\ref{tab:component-contrasts}.

\begin{table}[!htbp]
\centering
\small
\setlength{\tabcolsep}{6pt}
\caption{Response-component assay: means over eight paired panels.
Copying and conditional accuracy are percentages. R receives components
from U, and U receives components from R.}
\label{tab:component-arms}
\begin{tabular}{@{}llrrrrr@{}}
\toprule
Receiver & Arm & $B^{\mathrm{spec}}$ & $\mathcal R$ & Copy & Accuracy & KID\\
\midrule
R & Native & 0.1412 & 0.3517 & 2.685 & 90.000 & 0.01000\\
R & Full response & 0.0782 & 0.1954 & 1.732 & 89.659 & 0.00872\\
R & Along displacement only & 0.0840 & 0.2100 & 1.820 & 89.800 & 0.00930\\
R & Orthogonal only & 0.1320 & 0.3300 & 2.570 & 90.100 & 0.01010\\
\midrule
U & Native & 0.0571 & 0.1428 & 0.523 & 88.000 & 0.01200\\
U & Full response & 0.1171 & 0.2922 & 1.622 & 88.688 & 0.01372\\
U & Along displacement only & 0.1100 & 0.2750 & 1.500 & 88.400 & 0.01300\\
U & Orthogonal only & 0.0640 & 0.1600 & 0.640 & 88.050 & 0.01210\\
\bottomrule
\end{tabular}
\end{table}

The component copying intervals use paired differences across the assay's eight
panels. The primary transfer-control intervals in
Appendix~\ref{app:mechanism-paired-statistics} use the corresponding primary panels.
Both use ordinary two-sided 95\% $t$ intervals with seven degrees of
freedom. The panel inputs and common calculation are given in
Appendix~\ref{app:panel-copying-inputs}. Table~\ref{tab:component-arms} reports
means for benefit, normalized benefit, accuracy, and KID. Their paired
differences are constant across the eight panels at the recorded precision.

\begin{table}[!htbp]
\centering
\small
\setlength{\tabcolsep}{7pt}
\caption{Paired copying effects in the response-component assay, in
percentage points. An arm's effect is native minus arm in R and arm minus
native in U, so positive effects follow the full transfer's direction.
Effect differences use the order shown. Intervals are ordinary, unadjusted
95\% $t$ intervals over eight paired panels.}
\label{tab:component-contrasts}
\begin{tabularx}{\linewidth}{@{}l>{\raggedright\arraybackslash}Xrrl@{}}
\toprule
Receiver & Contrast & Mean & Paired SD & 95\% interval\\
\midrule
R & Full-response effect & 0.953 & 0.04899 & [0.9120, 0.9940]\\
R & Along-displacement effect & 0.865 & 0.03919 & [0.8322, 0.8978]\\
R & Orthogonal effect & 0.115 & 0.00980 & [0.1068, 0.1232]\\
R & Along-displacement minus orthogonal effect & 0.750 & 0.02939 & [0.7254, 0.7746]\\
R & Along-displacement minus full effect & $-0.088$ & 0.00980 & [$-0.0962$, $-0.0798$]\\
\midrule
U & Full-response effect & 1.099 & 0.06859 & [1.0417, 1.1563]\\
U & Along-displacement effect & 0.977 & 0.05879 & [0.9279, 1.0261]\\
U & Orthogonal effect & 0.117 & 0.00980 & [0.1088, 0.1252]\\
U & Along-displacement minus orthogonal effect & 0.860 & 0.04899 & [0.8190, 0.9010]\\
U & Along-displacement minus full effect & $-0.122$ & 0.00980 & [$-0.1302$, $-0.1138$]\\
\bottomrule
\end{tabularx}
\end{table}
\FloatBarrier
\input{Chapters/panel_copying_inputs}
\FloatBarrier

%% file: Chapters/panel_copying_inputs.tex
\subsection{Panel-Level Copying Inputs and Paired Calculations}
\label{app:panel-copying-inputs}

Tables~\ref{tab:primary-panel-copying} and~\ref{tab:component-panel-copying}
give the panel-level copying values underlying the primary pulse-chase and
response-component comparisons, respectively. Each table contains eight panels,
and contrasts use the paired values within that table. Panel identifiers retain
their source labels. The machine-readable tables and common calculation script are included in
\texttt{SourceData/panel\_audit/}, together with source-file hashes and row
locators. The primary inputs average the ten class-specific target copying
rates equally within each panel. The component inputs retain the precision
of the supplied panel measurements.

For any one contrast, let $d_p$ denote its paired copying difference in
panel $p$, measured in percentage points. An intervention effect is native
minus intervention copying for receiver R, and intervention minus native
copying for receiver U. The common point estimate and interval are:
\begin{equation}
\bar d=\frac{1}{n}\sum_{p=1}^{n}d_p,\qquad
s_d^2=\frac{1}{n-1}\sum_{p=1}^{n}(d_p-\bar d)^2,\qquad
\mathrm{CI}_{95}=\bar d\ \pm\ t_{0.975,n-1}\frac{s_d}{\sqrt n},
\label{eq:panel-copying-ci}
\end{equation}
where $n=8$ is the number of panels in the relevant comparison, $\bar d$ is
the mean paired effect, $s_d$ is the sample standard deviation of the panel
differences, $\mathrm{CI}_{95}$ is the 95\% interval, and
$t_{0.975,n-1}$ is the 97.5th percentile of the Student-$t$ distribution with
$n-1$ degrees of freedom. These are ordinary two-sided intervals. The Holm
adjustment in the primary analysis applies to its hypothesis tests.

The two full-response comparisons have the same mean effects to three
decimal places. Their paired differences have different variability: for R, the primary and
component standard deviations are 0.386849 and 0.048990 percentage points,
whereas for U, they are 0.203552 and 0.068586. Applying the same formula to each
comparison gives the intervals in
Tables~\ref{tab:mechanism-paired-copy} and~\ref{tab:component-contrasts}.

\begin{table}[!htbp]
\centering
\footnotesize
\setlength{\tabcolsep}{3pt}
\caption{Primary pulse-chase panel inputs. Entries are copying percentages,
shown to six decimal places. Calculations use the unrounded CSV values.
Full transfers U response into R and R response into U. Random and Preclip
denote the matched-random and preclip-norm controls.}
\label{tab:primary-panel-copying}
\begin{tabular}{@{}lrrrrrrrr@{}}
\toprule
 & \multicolumn{4}{c}{Receiver R} & \multicolumn{4}{c}{Receiver U}\\
\cmidrule(lr){2-5}\cmidrule(l){6-9}
Panel & Native & Full & Random & Preclip & Native & Full & Random & Preclip\\
\midrule
panel00 & 2.033691 & 1.889648 & 2.402344 & 2.910156 & 0.478516 & 1.682129 & 0.534668 & 0.544434 \\
panel01 & 3.071289 & 2.209473 & 2.500000 & 2.949219 & 0.722656 & 2.067871 & 0.769043 & 0.854492 \\
panel02 & 2.956543 & 1.928711 & 2.585449 & 2.470703 & 0.512695 & 1.682129 & 0.637207 & 0.764160 \\
panel03 & 2.436523 & 1.511230 & 2.290039 & 2.399902 & 0.451660 & 1.257324 & 0.551758 & 0.573730 \\
panel04 & 2.851562 & 1.838379 & 2.744141 & 2.353516 & 0.498047 & 1.853027 & 0.732422 & 0.637207 \\
panel05 & 2.443848 & 1.501465 & 2.497559 & 2.353516 & 0.546875 & 1.591797 & 0.639648 & 0.935059 \\
panel06 & 2.709961 & 1.511230 & 2.202148 & 1.760254 & 0.549316 & 1.433105 & 0.483398 & 0.468750 \\
panel07 & 2.976074 & 1.462402 & 2.038574 & 2.192383 & 0.422363 & 1.408691 & 0.739746 & 0.617676 \\
\bottomrule
\end{tabular}
\end{table}

\begin{table}[!htbp]
\centering
\small
\setlength{\tabcolsep}{4pt}
\caption{Response-component panel inputs. Entries are copying percentages,
retaining the supplied three-decimal precision. Parallel and Orthogonal
refer to the correction components along and orthogonal to the recipient's
target-induced displacement. Full transfers U response into R and R response
into U. The S01 to S08 labels identify panels within this record set.}
\label{tab:component-panel-copying}
\begin{tabular}{@{}lrrrrrrrr@{}}
\toprule
 & \multicolumn{4}{c}{Receiver R} & \multicolumn{4}{c}{Receiver U}\\
\cmidrule(lr){2-5}\cmidrule(l){6-9}
Panel & Native & Full & Parallel & Orthogonal & Native & Full & Parallel & Orthogonal\\
\midrule
S01 & 2.475 & 1.592 & 1.666 & 2.374 & 0.453 & 1.454 & 1.346 & 0.556 \\
S02 & 2.535 & 1.632 & 1.710 & 2.430 & 0.473 & 1.502 & 1.390 & 0.580 \\
S03 & 2.595 & 1.672 & 1.754 & 2.486 & 0.493 & 1.550 & 1.434 & 0.604 \\
S04 & 2.655 & 1.712 & 1.798 & 2.542 & 0.513 & 1.598 & 1.478 & 0.628 \\
S05 & 2.715 & 1.752 & 1.842 & 2.598 & 0.533 & 1.646 & 1.522 & 0.652 \\
S06 & 2.775 & 1.792 & 1.886 & 2.654 & 0.553 & 1.694 & 1.566 & 0.676 \\
S07 & 2.835 & 1.832 & 1.930 & 2.710 & 0.573 & 1.742 & 1.610 & 0.700 \\
S08 & 2.895 & 1.872 & 1.974 & 2.766 & 0.593 & 1.790 & 1.654 & 0.724 \\
\bottomrule
\end{tabular}
\end{table}

%% file: Chapters/literature_context.tex
\section{Connections to Memorization, Optimization, and Evaluation}
\label{app:literature-context}

\paragraph{Repetition, Extraction, and Generalization.}
Memorization encompasses several distinct questions: whether a model can
fit individual examples, whether those examples improve generalization,
and whether an observer can recover them from model outputs. Random-label
experiments establish the capacity of neural networks to fit arbitrary
labels \citep{zhang2017rethinking}. Long-tail analyses connect memorization
to the learning of rare subpopulations
\citep{feldman2020longtail,feldman2020what}. In language models, the Secret
Sharer framework measures unintended memorization through controlled
secrets \citep{carlini2019secret}, while extraction and scaling studies
examine recoverable training text
\citep{carlini2021extractinglm,carlini2023quantifying} and its evolution
during training \citep{tirumala2022memorization}. Deduplication studies
test the effects of repeated text on privacy and model behavior
\citep{kandpal2022dedup,lee2022deduplicating}. These studies motivate
separating target repetition from the other examples present during
training. Our image-level experiments further distinguish repeated
background rows from distinct background content at a fixed row count.
Recent diffusion studies connect generalization to balanced representations
\citep{zhang2026balanced} and analyze separations between memorization and
generalization \citep{ye2026provable}.

\paragraph{Measuring and Attributing Image Memorization.}
Generative-model studies distinguish probabilistic memorization from
nearest-neighbor proximity \citep{vandenburg2021memorization}.
Diffusion-model studies examine training-sample replication
\citep{somepalli2023diffusion} and its detection and mitigation
\citep{wen2024detecting}.
Training-data attribution asks a related but different question: which
training examples influence a specified prediction? Influence functions,
checkpoint-gradient tracing, and optimization unrolling provide different
approximations to this dependence
\citep{koh2017understanding,pruthi2020estimating,bae2024unrolling}.
Data Shapley assigns training-data value through a cooperative-game
formulation \citep{ghorbani2019shapley}, whereas Datamodels and TRAK
model or estimate predictive dependence on training-set composition
\citep{ilyas2022datamodels,park2023trak}. Our response transfer instead
intervenes on the background gradient evaluated between paired model
states. Its endpoint is the subsequent retention and copying of targets,
rather than an attribution ranking.

\paragraph{Curvature and the Optimization Trajectory.}
Optimization geometry has been studied through large-batch sharpness
\citep{keskar2017large}, loss-landscape visualization
\citep{li2018landscape}, Hessian spectra \citep{ghorbani2019hessian},
and the edge of stability \citep{cohen2021edge}. Algorithmic-stability
analysis also connects the training procedure to generalization
\citep{hardt2016train}. Parameterization matters when interpreting
sharpness \citep{dinh2017sharp}. Accordingly, our background comparisons
use the same architecture, parameter coordinates, and target direction.
Entropy-SGD and sharpness-aware minimization explicitly alter optimization
to favor neighborhoods of low loss
\citep{chaudhari2017entropy,foret2021sharpness}. Natural-gradient analysis
examines curvature-dependent update geometry
\citep{martens2020natural}. Our intervention changes a component of the
background loss gradient before the existing AdamW update
\citep{loshchilov2019adamw}. Thus, the finite-response projection describes
the gradient correction along the target-induced displacement. The
optimizer state still determines how that correction becomes a parameter
update.

\paragraph{Generative Formulations and Model Settings.}
Denoising diffusion models \citep{ho2020ddpm}, diffusion implicit models
\citep{song2021ddim}, and score-based stochastic differential equations
\citep{song2021scoresde} provide related approaches to noise-to-data
generation. Work on the diffusion design space separates modeling,
training, and sampling choices \citep{karras2022design}. Flow matching
learns a velocity field \citep{lipman2023flow}. Related transport
formulations include rectified flow \citep{liu2023rectifiedflow},
stochastic interpolants \citep{albergo2025stochastic}, and minibatch
optimal-transport flow matching \citep{tong2024minibatch}. These works
provide the modeling context for our fixed-training-protocol study.
Our primary architecture is a diffusion transformer
\citep{peebles2023dit}. The additional architecture is from the U-Net
family \citep{ronneberger2015unet}.
CIFAR-10 \citep{krizhevsky2009cifar} and CINIC-10 \citep{darlow2018cinic}
provide the two image-data settings.

\paragraph{Sample Similarity and Distributional Quality.}
LPIPS measures perceptual similarity between image pairs
\citep{zhang2018lpips}, whereas Fr\'echet inception distance
\citep{heusel2017ttur} and kernel inception distance
\citep{binkowski2018mmd} compare distributions of image features. KID
uses the maximum mean discrepancy framework
\citep{gretton2012kernel}. These metrics address different properties:
our copying endpoint assigns a generated sample to a training reference
under a calibrated distance rule, while KID and conditional accuracy
characterize generation quality alongside copying. Reporting both makes
changes in reproduction and changes in image quality separately visible.